\documentclass[a4paper,10pt,twocolumn]{article}

\usepackage{sectsty}
\allsectionsfont{\sffamily}
\usepackage{parskip}

\usepackage{cite}

\usepackage[top=2.0cm,bottom=2.0cm,left=2.0cm,right=2.0cm]{geometry}

\usepackage{amsmath,amssymb}

\usepackage{graphicx}
\usepackage[hang,tight,footnotesize]{subfigure}

\usepackage{tikz}
\begin{document}

%\title{\bfseries \sffamily Partial Isometric Correspondence by Local Metric Matching}
\title{\bfseries \sffamily A Low-Dimensional Representation for Robust Partial Isometric Correspondences Computation}

\author{Alan Brunton\thanks{Max Planck Institute for Informatics, Germany, \{abrunton,mwand,weinkauf,hpseidel\}@mpi-inf.mpg.de} \and Michael Wand\footnotemark[1] \and Stefanie Wuhrer\thanks{Cluster of Excellence, Multi-Modal Computing and Interaction, Saarland University, Germany, swuhrer@mmci.uni-saarland.de} \and Hans-Peter Seidel\footnotemark[1] \and Tino Weinkauf\footnotemark[1]}

\date{}

% comments
\definecolor{checkmecol}{rgb}{0.7,0.2,0.3}
\newcommand{\xxx}{{\color{checkmecol} \ttfamily \small \textbf{xxx}}}
\newcommand{\remark}[1] {{\color{checkmecol} \ttfamily \small \textbf{[#1]}}}

% formatting
\newcommand{\argmax}{\operatorname{argmax}}
\newcommand{\argmin}{\operatorname{argmin}}
\newcommand{\dist}{\operatorname{dist}}
\newcommand{\reals}{\mathbb{R}}
\newcommand{\integers}{\mathbb{Z}}
\newcommand{\bv}[1]{\mathbf{#1}}

% notation
\newcommand{\surfaceS}{\mathcal{S}}
\newcommand{\surfaceT}{\mathcal{T}}
\newcommand{\vecx}{\bv{x}}
\newcommand{\vecy}{\bv{y}}
\newcommand{\vecn}{\bv{n}}
\newcommand{\vecu}{\bv{u}}
\newcommand{\vecv}{\bv{v}}
\newcommand{\vecs}{\bv{s}}
\newcommand{\vect}{\bv{t}}
\newcommand{\vecd}{\bv{d}}
\newcommand{\vecD}{\bv{D}}

\newcommand{\firstFund}{\mathbb{I}}

%%%%%%%%%%%%%%%%%%%%%%%%%%%%%%%%%%%%%%%%%%%%%%%%%%%%%%%%%%%%%
%% OPTIONS FOR SUBFIGURES
%%%%%%%%%%%%%%%%%%%%%%%%%%%%%%%%%%%%%%%%%%%%%%%%%%%%%%%%%%%%%
%%Redefinition of Labels for Subfigures
%%Example:
%%In the figure 4.1 simply: (a)
%%As Reference inside the text: 4.1a
\makeatletter
    \renewcommand{\thesubfigure}{\alph{subfigure}}
    \renewcommand{\@thesubfigure}{\subcaplabelfont (\thesubfigure)\space}
    \renewcommand{\p@subfigure}{\thefigure}
\makeatother

%%Definitions for spaces between the Subfigures
%% The Commands '\goodgap' and '\littlegap' are defined here.
\newlength{\lengthgoodgap}
\addtolength{\lengthgoodgap}{\subfigtopskip}
\addtolength{\lengthgoodgap}{\subfigbottomskip}
\newcommand{\goodgap}{\hspace{\lengthgoodgap}}
\newlength{\lengthlittlegap}
\addtolength{\lengthlittlegap}{\subfigtopskip}
\newcommand{\littlegap}{\hspace{\lengthlittlegap}}

%%Definition for size of pictures, if 2 in one row
\newlength{\twopicwidth}
\addtolength{\twopicwidth}{0.5\textwidth}
\addtolength{\twopicwidth}{-0.5\lengthgoodgap}

%%Definition for size of pictures, if 3 in one row
\newlength{\threepicwidth}
\addtolength{\threepicwidth}{0.333333\textwidth}
\addtolength{\threepicwidth}{-0.666666\lengthlittlegap}

%%Definition for size of pictures, if 4 in one row
\newlength{\fourpicwidth}
\addtolength{\fourpicwidth}{0.25\textwidth}
\addtolength{\fourpicwidth}{-0.75\lengthgoodgap}

%%Definition for size of pictures, if 5 in one row
\newlength{\fivepicwidth}
\addtolength{\fivepicwidth}{0.20\textwidth}
\addtolength{\fivepicwidth}{-0.80\lengthlittlegap}

%%Definition for size of pictures, if 6 in one row
\newlength{\sixpicwidth}
\addtolength{\sixpicwidth}{0.166666666666666667\textwidth}
\addtolength{\sixpicwidth}{-0.833333333333333333\lengthlittlegap}

%%For text with columns - Definition for size of pictures, if 2 in one row
\newlength{\ltwopicwidth}
\addtolength{\ltwopicwidth}{0.5\linewidth}
\addtolength{\ltwopicwidth}{-0.5\lengthgoodgap}

%%For text with columns - Definition for size of pictures, if 3 in one row
\newlength{\lthreepicwidth}
\addtolength{\lthreepicwidth}{0.333333\linewidth}
\addtolength{\lthreepicwidth}{-0.666666\lengthlittlegap}

%%For text with columns - Definition for size of pictures, if 4 in one row
\newlength{\lfourpicwidth}
\addtolength{\lfourpicwidth}{0.25\linewidth}
\addtolength{\lfourpicwidth}{-0.75\lengthlittlegap}

%%%Kill the space at top of the subfigures - BE CAREFULL HERE!
%%%Basically, this is a problem with my beloved hyperref:
%%% The value of \lastskip seems to get destroyed and therefore the subfigure code breaks.
%\ifx\hypersetup\undefined
	%\newlength{\manualsubfigtopskip}
	%\addtolength{\manualsubfigtopskip}{0cm}
%\else
	%\newlength{\manualsubfigtopskip}
	%\addtolength{\manualsubfigtopskip}{\subfigtopskip}
	%\setlength{\subfigtopskip}{0cm}
%\fi
%

%%%%%%%%%%%%%%%%%%%%%%%%%%%%%%%%%%%%%%%%%%%%%%%%%%%%%%%%%%%%%
%% TikZ DEFINITIONS
%%%%%%%%%%%%%%%%%%%%%%%%%%%%%%%%%%%%%%%%%%%%%%%%%%%%%%%%%%%%%

\tikzset{KayleighArrow/.style={-latex, cap=rect, draw=black, fill=none, thick, inner sep=0pt}}%

\newcommand{\KayGrid}{
	\draw[style=help lines, color=gray, xstep=0.05, ystep=0.05] (0,0) grid (1, 1);%
	\draw[style=help lines, color=darkgray, xstep=0.1, ystep=0.1, thick] (0,0) grid (1, 1);%
	\draw[style=help lines, color=black, xstep=0.5, ystep=0.5, very thick] (0,0) grid (1, 1);%
}

\newcommand{\KayGridHalf}{
	\draw[style=help lines, color=gray, xstep=0.05, ystep=0.05] (0,0) grid (1, 0.5);%
	\draw[style=help lines, color=darkgray, xstep=0.1, ystep=0.1, thick] (0,0) grid (1, 0.5);%
	\draw[style=help lines, color=black, xstep=0.5, ystep=0.5, very thick] (0,0) grid (1, 0.5);%
}

%%\unitlen not defined here
%\tikzset{KayleighText/.style={fill=white, inner sep=0.01\unitlen}}%

%%%%%%%%%%%%%%%%%%%%%%%%%%%%%%%%%%%%%%%%%%%%%%%%%%%%%%%%%%%%%
%% OTHER OPTIONS AND DEFINITIONS
%%%%%%%%%%%%%%%%%%%%%%%%%%%%%%%%%%%%%%%%%%%%%%%%%%%%%%%%%%%%%
%%LaTeX shall not issue a warning, if Overfull Box < 3pt
\hfuzz=3pt

%%%%%%%%%%%%%%%%%%%%%%%%%%%%%%%%%%%%%%%%%%%%%%%%%%%%%%%%%%%%%
%% Resetting a LaTeX length
%%%%%%%%%%%%%%%%%%%%%%%%%%%%%%%%%%%%%%%%%%%%%%%%%%%%%%%%%%%%%
%Be aware that there shall be no space in '\ifx#1\undefined'
\newcommand{\resetlength}[1]{\ifx#1\undefined \newlength{#1} \else \setlength{#1}{0pt} \fi}

%%%%%%%%%%%%%%%%%%%%%%%%%%%%%%%%%%%%%%%%%%%%%%%%%%%%%%%%%%%%%%
%%% Patch some hyperref stuff
%%%%%%%%%%%%%%%%%%%%%%%%%%%%%%%%%%%%%%%%%%%%%%%%%%%%%%%%%%%%%%
%%\ifx\texorpdfstring\undefined
%%
%%	\newcommand{\texorpdfstring}[2]{#1}
%%
%%\fi
%
%
%
%%\newlength{\abovecaptionskip}
%%\newlength{\belowcaptionskip}
%%\setlength{\abovecaptionskip}{0pt}
%%\setlength{\belowcaptionskip}{0pt}
%%\newcommand\figurename{Figure}
%
%%For Computer Graphics styles, i.e., EuroGraphics, EuroVis:
%%\parskip 5.5pt plus 1pt minus 1.5pt
%\parskip 5pt plus 1pt minus 1.5pt
%

%\graphicspath{{pdfimages/}}

\maketitle

\begin{abstract}
Intrinsic shape matching has become the standard approach for pose invariant correspondence estimation among deformable shapes. Most existing approaches assume global consistency. While global isometric matching is well understood, only a few heuristic solutions are known for partial matching. Partial matching is particularly important for robustness to topological noise, which is a common problem in real-world scanner data.
We introduce a new approach to partial isometric matching based on the observation that isometries are fully determined by local information: a map of a single point and its tangent space fixes an isometry. We develop a new representation for partial isometric maps based on equivalence classes of correspondences between pairs of points and their tangent-spaces. We apply our approach to register partial point clouds and compare it to the state-of-the-art methods, where we obtain significant improvements over global methods for real-world data and stronger guarantees than previous partial matching algorithms.
\end{abstract}

\section{Introduction}

Modern computer graphics has experienced a paradigm shift: Traditional manual modeling is increasingly complemented by data-driven techniques where measured data, such as 3D scans, are used as a basis for building 3D models. 
An important data source are 3D scans of deformable models, such as humans or animals in varying poses. Today, there exist sophisticated scanning setups for acquiring moving geometry in real-time \cite{Davis2005,Konig2007,Weise2007,Vlasic2009} and there are even consumer devices on the market such as the Microsoft Kinect$^{\text{TM}}$. This leads to new applications such as virtual cinematography \cite{Debevec2006}, or the creation of data-driven generative shape models of deformable objects \cite{Blanz1999,Allen2003,Hasler2009,Weise2011}. Finding correspondences among such data is a fundamental problem for all of these applications: Almost any further processing, such as the registration of partial scans into a complete shape, editing of sequences, or statistical analysis, requires dense correspondences between surface points.

Matching deformable shapes is in many cases a difficult problem: If we permit rather general deformations this might require many parameters that have to be explored during matching. The size of this search space is exponential with respect to the available degrees of freedom. However, for the important special case of a single object in different poses, we can often assume that the deformation is approximately isometric, i.e., preserving the intrinsic metric structure. Concretely, the distances along surfaces of objects such as humans, animals, plants, or cloths do not change a lot without serious injury or damage. This restriction leads to a strongly constrained search space. Lipman and Funkhouser~\cite{Lipman2009} argue that isometries between topological disks are a special case of conformal mappings, thereby limiting the degrees of freedom to six (three point-to-point correspondences are sufficient). Ovsjanikov et al.~\cite{Ovsjanikov2010} show that a single point correspondence is sufficient for a special class of shapes where the spectrum of the Laplace-Beltrami operator is not degenerate, thus showing that there are only two degrees of freedom in this special case.
Additional cues, such as carefully selected constellations of local features \cite{Tevs2011}, can even reduce the complexity for many shapes to the point of leaving a trivial search space, eliminating all degrees of freedom.
As approximate isometry makes the correspondence problem feasible while still permitting significant pose changes, many of the recent shape matching algorithms are based on this assumption \cite{Anguelov2004,Bronstein2006,Huang2008,Tevs2009,Sun2009,Lipman2009,Ovsjanikov2010,Kim2011,Tevs2011,Ovsjanikov2012}.

However, the isometric matching problem is not yet solved: Because of the intrinsic view of the geometry, it is naturally sensitive to topological noise. In case of holes, missing data, or contacts, intrinsic distances become distorted and thus no longer constitute invariants that can be exploited for matching. One solution is to replace the notion of distance. For example, by using diffusion distances, or variants thereof, one can reduce the sensitivity to topological artifacts \cite{Sun2009,Bronstein2010,Lipman2010Biharmonic} when the pieces of geometry that cause the problem are small in relation to the overall shape. Nonetheless, these invariants still break down in case of large artifacts (wide contacts, large holes, as shown for example in Section~\ref{sec:results} of this paper). Unfortunately, real-world 3D scanner data, one of the main practical application areas, is almost universally troubled by substantial artifacts of this kind.

Formulated more generally, we have to address the problem of \emph{partial matching}, where not the whole manifold can be brought into correspondence by a (near-) isometric mapping but only an excerpt of the surface can be matched. In this case, typical invariants (geodesic paths, Laplacian eigenfunctions) become unreliable because they utilize global information. Importantly, the excerpts of the surfaces that can be matched are not known a priori (otherwise, we could just restrict a traditional method accordingly) but need to be determined along with the matching. This seems to re-introduce prohibitive complexity as we now have to choose from an exponential number of such subsets \cite{Xu2012}. The most important contribution of this paper is to show that with an appropriate matching model this is not the case. The search space is not much larger than in the global problem and we give an efficient algorithm for computing such matches.

The core of our method is based on the observation from differential geometry that an isometric map can be fully specified by using local information up to first order only: An isometry between Riemannian manifolds is fixed by a single point correspondence and an orthogonal map of the tangent spaces (see for example~\cite[p. 201]{Berger2002}). In the case of surfaces, this means that the map of a point and a local direction (plus orientation, in case of unoriented surfaces) is sufficient to determine an isometry. 
We will sketch a constructive proof in Section~\ref{sec:proposed} that directly yields a propagation algorithm for computing matches: Starting with a single oriented point match, correspondence information is incrementally propagated to the neighborhood, thereby flood-filling a partially consistent region of isometric geometry. Being local in nature, the method handles partial matches naturally and is robust to topological noise, which is reported naturally as boundary of partiality.

From a structural point of view, we can understand partial isometries among smooth, connected manifolds as equivalence classes of mappings of a pair of points and their tangent spaces on the two surfaces involved. In the case of oriented 2-manifolds, each such object has six degrees of freedom (a point and an angle around the surface normal for each for source and target surface, respectively). The mapping is captured by a equivalence class of such 6-tuples. This set is redundant in that one degree of freedom (the sum of the angles) can be chosen arbitrarily and two further parameters (either the starting or the end point) are only needed to select the partial region to be mapped. This leaves three degrees of freedom that need to be actually explored densely, with false starting points being rejected in $\mathcal{O}(1)$ time.% In practice, we can therefore expect costs cubic in the intended accuracy. Higher costs up to fifth order can only occur in highly symmetric shapes (for example matching two spheres), were many overlapping correct matches exist which are enumerated in an output-sensitive fashion.} %computationally.

In this view, we can perform a relaxation: In order to also find \emph{approximate} isometries, we can cast this problem as the task of finding \emph{approximate equivalence classes}. Kim et al.~\cite{Kim2011} have used this idea in the context of globally consistent isometric mappings in order to efficiently find approximate isometries. In our paper, we demonstrate how our partial matching framework can be adapted to perform the same task in a partial matching scenario. We perform agglomerative clustering~\cite{zhang_gdl_eccv2012} in the space of near-isometric mappings, which are concisely represented using the tuples introduce above.

We validate our algorithm on standard benchmark data sets and raw scanner data, and compare the results to previous work. We show a significant improvement in quality over global methods in shapes with topological noise. Our algorithm yields similar or better results as previous heuristics for partial matching, but with stronger guarantees of discovering existing isometries as outlined above.

In summary, our paper proposes a systematic framework and new algorithms for extending isometric matching to the case of partial consistency, thereby making the following specific contributions:

\begin{itemize}
	\item We characterize partial isometries of shapes by single-point maps up to first order, which yields a tight bound on the inherent degrees of freedom.
	\item This leads to a novel matching algorithm that provides a systematic approach to the general setting of partial intrinsic matching, where both surfaces may be incomplete, including robustness to strong topological noise.
	\item By interpreting partial matching as a problem of finding approximate equivalent classes in our novel representation, we obtain an algorithm for \emph{approximate} partial isometric matching.
\end{itemize}

Our algorithmic pipeline for approximate partial isometric matching is summarized as follows. We identify distinctive feature points on both surfaces. We compute oriented point correspondences by matching feature descriptors, with orientation determined by nearby features. From oriented point matches, we perform local metric propagation, stopping the propagation when the stretching becomes too large. This gives us a set of partial isometries, covering different, but possibly overlapping regions of both surfaces. We define a dissimilarity measure between partial isometries, and use this to cluster the partial matches into equivalence classes. The cluster (equivalence class) with the smallest total intra-cluster distance is merged by taking the geodesic centroid of all candidate correspondences for each point on the source manifold.

The steps in this pipeline exhibit sensitivity to certain challenges. Most prominent are the sensitivity of feature matching to surface noise, missing data and topological noise, and the sensitivity of the clustering step to under-sampling of the space of isometric mappings, which can be exacerbated by problems in the feature matching stage, in addition to the lack of a proper distance metric between partial matches. The difficulty of reliable feature matching on real data can be alleviated by the metric propagation algorithm, which will typically produce smaller partial isometries for incorrect feature matches than for correct ones since we expect the local metric to be vary significantly for different parts of the surface. However, in the presence of strong continuous intrinsic symmetries this assumption breaks down. We thoroughly discuss and explore in which situations and to what extent these challenges create problems in the final result in Section \ref{sec:results}.

%The remainder of the paper is structured as follows: In Section~\ref{sec:what-is-difficult}, we review different matching strategies in literature and classify them by complexity of the arising search space. In Section~\ref{sec:proposed}, we introduce and motivate our matching model, analyze the degrees of freedom, and develop the local propagation algorithm for computing near-isometric matches. In order to find starting points of the propagation algorithm, we use conventional feature matching. This is not a contribution of this paper, but we briefly describe the solution employed in our experiments in Section~\ref{sec:pairwise}. The section also introduces our clustering heuristic for handling approximate partial isometry where the deviations from exact isometry are substantial. Finally, Section~\ref{sec:results} performs an empirical evaluation of the proposed techniques and Section~\ref{sec:conclusions} concludes with some ideas for future work.

\section{Complexity of Isometric Shape Matching}
\label{sec:what-is-difficult}

In this section, we discuss different isometric matching models and their implications on the difficulty for finding a globally optimal solution to the matching problem. To the best of our knowledge, this has not yet been analyzed explicitly in literature. We do not aim to give an extensive review of surface registration methods; for this, we refer the reader to recent surveys such as van Kaick et al.~\cite{vanKaick_survey_extended_2011} or Tam et al.~\cite{tam_survey_2013}. 

We start by introducing some formal notations.

\noindent\textbf{Manifolds:} We consider smooth, orientable $2$-manifolds $\mathcal{M}\subset \reals^3$ embedded in three-dimensional space. In order to represent partial data (such as scans with acquisition holes), we permit boundaries, denoted by $\partial \mathcal{M}$. The orientation of $\mathcal{M}$ might (optionally) be prescribed by oriented surface normals $\vecn(\vecx)$, $\vecx \in \mathcal{M}$, pointing outwards.

\noindent\textbf{Tangent space:} We further use $T_\vecx\mathcal{M}$ to denote the tangent space of $\mathcal{M}$ at point $\vecx \in \mathcal{M}$. For its representation, we choose two arbitrary but fixed orthogonal tangent vectors $\vecu(\vecx),\vecv(\vecx)$, i.e.: $T_\vecx\mathcal{M} = \operatorname{span}\{\vecu(\vecx),\vecv(\vecx)\}$. %The choice of a fixed coordinate frame is also denoted as a matrix by $\bv{T}(\vecx) = \left(\vecu(\vecx)|\vecv(\vecx)\right)$.

\noindent\textbf{Distances:} We use $\dist_\mathcal{M}(\vecx,\vecy)$ for $\vecx,\vecy \in \mathcal{M}$ to denote the \emph{intrinsic} or \emph{geodesic distance} between the two points $\vecx$ and $\vecy$. A \emph{geodesic} connecting $\vecx$ and $\vecy$ is a curve that has no geodesic curvature, which means that the derivative of the curve at a point $\vecs$ projected to $T_\vecs\mathcal{M}$ is zero. We call the shortest geodesic connecting $\vecx$ and $\vecy$ a \emph{shortest geodesic path} in $\mathcal{M}$.

\noindent\textbf{Mappings and isometries:}
Consider two manifolds $\surfaceS$ and $\surfaceT$, and a mapping $f: \mathcal{U} \rightarrow \surfaceT$ from an open subset $\mathcal{U} \subseteq \surfaceS$ to $\surfaceT$. Let $f_{\vecu}(\vecx)$ and $f_{\vecv}(\vecx)$ denote the partial derivatives of $f$ with respect to the tangent space directions $\vecu$ and $\vecv$ in $\reals^3$ of $\surfaceS$.
The first fundamental form $\firstFund_f(\vecx)$ of $f$ at point $\vecx \in \mathcal{U}$ is then given by:
\[
\firstFund_f(\vecx) = \left( \begin{matrix}
f_{\vecu}(\vecx) \cdot f_{\vecu}(\vecx) & f_{\vecu}(\vecx) \cdot f_{\vecv}(\vecx) \\
f_{\vecv}(\vecx) \cdot f_{\vecu}(\vecx) & f_{\vecv}(\vecx) \cdot f_{\vecv}(\vecx)
\end{matrix} \right).
\]
The function $f$ is an \emph{isometry} if and only if $\firstFund_f(\vecx) = \bv{I}$ for all $\vecx \in \mathcal{U}$. 

Equipped with this notation, we will now identify and analyze three classes of approaches for finding surface correspondences. The difference is in how the notion of \emph{approximate} isometry is handled, leading to different complexity characteristics and algorithms.

\subsection{Global Approximate Isometry}

The global model assumes that \emph{geodesic distances} are \emph{global invariants} of the shape, being consistent at least up to an error margin $\nu > 0$ that accounts only for a small fraction of the object size. This means, the energy
\begin{equation}
E_{global} = \sup_{\vecx,\vecy \in \mathcal{U}} | \dist_\mathcal{S}(\vecx,\vecy) - \dist_\mathcal{T}(f(\vecx),f(\vecy)) |
\label{eq:globalApproxIso}
\end{equation}
must be smaller than $\nu$. For two points $\vecx,\vecy \in \mathcal{U}$ Eq.~(\ref{eq:globalApproxIso}) corresponds to an additive error of at most $\nu$, i.e.
\begin{equation}
| \dist_\mathcal{S}(\vecx,\vecy) - \dist_\mathcal{T}(f(\vecx),f(\vecy)) | \leq \nu.
\label{eq:absoluteError}
\end{equation}
The global consistency criterion is sometimes modified to allow for a multiplicative error of $\nu$ instead, as
{\small
\begin{equation}
\max \left( \frac{\dist_\mathcal{S}(\vecx,\vecy)}{\dist_\mathcal{T}(f(\vecx),f(\vecy))}, \frac{\dist_\mathcal{T}(f(\vecx),f(\vecy))}{\dist_\mathcal{S}(\vecx,\vecy)} \right) \leq (1 + \nu).
\label{eq:relativeError}
\end{equation}
}
The former error model considers absolute errors, while the latter one considers relative errors.

In case of exact isometry, i.e., $\nu=0$, the set of matching candidates becomes strongly constrained. The isometry assumption has been used to embed the intrinsic geometry of a shape in a Euclidean space using multi-dimensional scaling, such that embeddings of isometric shapes become identical, which facilitates shape recognition and matching~\cite{Elad2003,jain_zhang_kaick_06_nonrigid_spectral_correspondence,Wuhrer2007}. Alternatively, the geometry of shape $\mathcal{S}$ can be embedded into shape $\mathcal{T}$ using generalized multi-dimensional scaling, thereby computing a cross-parameterization directly~\cite{Bronstein2006}.

Exact isometry results in a set of matching candidates with few degrees of freedom.
Lipman and Funkhouser~\cite{Lipman2009} have noticed that isometries are special cases of conformal maps, thereby having only the degrees of freedom given by the M\"{o}bius group, which are fixed by three pointwise matches on spherical topologies.
Ovsjanikov et al.~\cite{Ovsjanikov2010} have shown that even a single point match is sufficient to fix an isometry if the Laplace-Beltrami spectrum of $\mathcal{S}$ is non-degenerate.
In this paper,
we exploit a different way to uniquely describe an isometry:
fixing one point, a tangential direction, and the surface orientation is necessary and sufficient to specify an isometric mapping~\cite{Berger2002}.
This provides the fewest possible degrees of freedom
while still covering all cases including shapes with global intrinsic symmetries.

In case of small, global error margins, statistical triangulation algorithms can be applied that compute all correspondences from a few landmark matches \cite{Huang2008,Tevs2009} or, similarly, by voting for several approximate solutions \cite{Lipman2009,Ovsjanikov2010}. Depending on the geometry of the shape, errors can become amplified so that a bit more than only the minimal set of initial correspondences are required \cite{Tevs2011}. Nonetheless, matching according to this model is efficient and can be considered a more or less solved problem. 

The global approximate isometry criterion has also been employed to study matching partial overlaps of complete surfaces. van Kaick et al.~\cite{vankaick2013_bilatmap} use a pair of features to define a map that captures a local geodesic region between the two features, and show how this descriptor can be used for partial matching. Here, using two features instead of one is crucial because two features encode orientation information on the surface, while one feature does not.

The drawback of the global consistency model is its sensitivity to topological noise.
To make globally consistent approaches more robust w.r.t.\ topological changes, several approaches have been proposed to perform a band-limited analysis in the eigenspace of the manifold's Laplace-Beltrami operator~\cite{Sun2009,Bronstein2010,Lipman2010Biharmonic,Aubry2011}. Unlike prior approaches based on embedding the intrinsic geometry of the shape directly~\cite{Elad2003,jain_zhang_kaick_06_nonrigid_spectral_correspondence,Wuhrer2007,Bronstein2006}, these approaches successfully handle small topological errors. However, they break down in the presence of large artifacts, such as wide contacts or missing pieces. A notable exception is the heuristic region growing approach by Sharma et al.~\cite{Sharma2011}. It tries to match points with similar spectral signatures using an expectation--maximization framework, which has been shown to perform well in the presence of large contacts. 
Despite good practical performance, the algorithm is heuristic in nature and it remains unclear under which conditions it will find a correct solution. In particular, if local descriptors are not unique, the greedy region growing might fail and the EM-algorithm does not guarantee to recover a correct global match. In contrast, our region growing algorithm, which is based on propagation of metric information rather than potentially ambiguous descriptor matching, comes with the theoretical guarantee to find correct results for exact isometries while requiring an initialization from a small search space with very few degrees of freedom.

\subsection{Global Approximate Isometry in Partial Regions}

The global consistency model is incompatible with the notion of partial matching, since distances have to be measured on the complete surfaces $\mathcal{S}$ and $\mathcal{T}$, which might not be available.  Xu et al.~\cite{Xu2012} modify the criterion in Eq.~(\ref{eq:globalApproxIso}) by restricting paths to the partially matched region $\mathcal{U}$:
\begin{equation}
E_{partial} = \sup_{\vecx,\vecy \in \mathcal{U}} | \dist_\mathcal{U}(\vecx,\vecy) - \dist_{f(\mathcal{U})}(f(\vecx),f(\vecy)) |.
\label{eq:globalApproxIsoXu}
\end{equation}
Note that $E_{partial}$ considers again absolute errors
\begin{equation}
| \dist_\mathcal{U}(\vecx,\vecy) - \dist_{f(\mathcal{U})}(f(\vecx),f(\vecy))| \leq \nu
\label{eq:absoluteErrorPartial}
\end{equation}
and can be modified to consider relative errors as in Eq.~(\ref{eq:relativeError}).

The drawback is that the shortest geodesic paths and thus the energy depend on the shape of the domain, which makes it difficult to optimize $E_{partial}$; changing the domain $\mathcal{U}$ influences which pairs of points are mapped in a geodesically consistent way. For this reason, Xu et al. restrict their method to considering geodesically convex regions $\mathcal{U}$, which are defined as regions where for any two points $\vecx, \vecy$ in $\mathcal{U}$, there exists a shortest geodesic path between $\vecx$ and $\vecy$ in $\mathcal{U}$. In this special case, both $E_{partial}$ and $E_{global}$ measure shortest paths on $\surfaceS$ and $\surfaceT$. 
The solution proposed by Xu et al. optimizes the scale of consistent regions and the consistent points separately, which leads to a rather complex algorithm. Further, the notion of a scale parameter is not canonically related to the original matching problem.

Sahilloglu and Yemez~\cite{sahillioglu_pg2012} consider the case where one of the surfaces is complete and the other an incomplete, deformed part of that surface. Using a coarse sampling and matching strategy between shape extremities, they can directly estimate a scale parameter between the two surfaces, which allows them to define a scale-invariant isometric distortion measure. This results in one-sided partial dense intrinsic matching up to an arbitrary scale. Their approach also allows matching of semantically similar, but non-isometric complete surfaces. Our approach allows both surfaces to be incomplete, at the cost that they must be scaled consistently beforehand. Given real-world scanners often provide data in known units, our method is compatible with the scenario of matching surfaces acquired with different modalities. While in this paper we do not explore the latter scenario, our approach would be compatible with this task given a reliable way to estimate scale change during metric propagation, possibly using shape extremities or other intrinsic features.

Bronstein et al.~\cite{bronstein_etal_partial_sim_ijcv_2009} introduced a general framework to evaluate partial similarity using Pareto optimality. In case of partial intrinsic shape matching, this method aims to find large parts of two surfaces that are similar to each other, where similarity is measured according to Eq.~(\ref{eq:absoluteError}). In practice, the parts are found using generalized multi-dimensional scaling. Raviv et al.~\cite{raviv_etal_partial_sym_ijcv_2010} use a similar technique to find partial intrinsic symmetries. 

\subsection{Local Approximate Isometry}

Another common way to relax the requirement of exact isometry towards approximate matching is to maintain the metric tensor in a least-squares-sense. Again, let $f: \mathcal{U} \rightarrow \surfaceT, \mathcal{U} \subseteq \surfaceS$ be a mapping between two manifolds, where $\mathcal{U}$ is the open subset of $\surfaceS$ that should be mapped to $\mathcal{T}$ in a distortion minimizing way. We can measure the distortions for example by minimizing a matrix norm of the Green deformation tensor (difference of the first fundamental form to identity):
\begin{equation}
E_{local}(f) = \int_{\mathcal{U}} \left\| \firstFund_f(\vecx) - \bv{I} \right\|^2_F d \vecx.
\label{eq:localApproxIso}
\end{equation}
This criterion is purely local and thus well suited for partial matching. It is worth noting that extrinsic elastic deformation techniques, such as \cite{Haehnel2003,Zhang2008,Li2008} are closely related: They either include the preservation of the curvature tensor in the objective function to maintain the extrinsic shape, or apply Eq.~(\ref{eq:localApproxIso}) to the 3-manifold of the embedding Euclidean volume \cite{Wand2009,LiEtAl2009}. All of these methods are designed for partial matching.

The problem with both intrinsic and extrinsic elastic matching models is that the search space becomes very large, rendering any approach based on exhaustive search prohibitively expensive. 
The structure of the search space can be approximately understood by a linearized analysis. In order to understand the degrees of freedom of the local matching model, we can use the tool of modal analysis of such elastic models, first introduced by Pentland and Williams~\cite{Pentland89} to the field of computer graphics (the aim of their paper was actually to speed up the simulation of extrinsic elastic deformations of solid objects in three-space). 
Modal analysis represent the deformation of an object as a linear combination of the eigenmodes of the object's vibration, which are found using a spectral analysis of a linearized deformation energy. 

Typically, these energies are related to the Laplacian of the deformed domain, thus leading to eigenvalues that are only decreasing rather slowly. Many modes need to be retained to represent the space of low-energy (i.e., plausible) deformations adequately. If the local matching is not very stiff, the size of the search space explodes. This is intuitive: Permitting local deformations creates a large variety of permissible shape variants, for example by adding different local dents and combinations of those everywhere. % \cite{Pentland89}. %(due to Gibbs phenomenon) 
Due to this large search space, it is impractical to match shapes purely based on the deformation model of Eq.~(\ref{eq:localApproxIso}). In order to reduce the large search space for elastic matching, existing methods use additional constraints, such as, template models~\cite{LiEtAl2009}, temporal coherence~\cite{Wand2009}, or fairly large sets of coherent feature correspondences~\cite{Zhang2008}.

Recently, Kim et al.~\cite{Kim2011} proposed a new paradigm for the elastic matching problem. First, the method computes multiple dense maps between two shapes by assuming a global near-isometric deformation model. Multiple maps are obtained by fixing different triples of corresponding points for the computation of global conformal maps~\cite{Lipman2009}. Second, the method computes local weights for pairs of maps, depending on their local adherence to isometry, and performs a spectral analysis to combine multiple weighted maps into a single global map. The key idea is to find cliques of similar mappings by clustering approximately compatible maps.
The method was shown to perform well in many interesting cases.
However, it cannot handle partial mapping; in particular constellations with large topological noise cannot be handled: Each global conformal map can is highly distorted due to the lack of global consistency. This introduces distortions in many of the local matches, and it is not always possible to remove these distortions in the final blended result (as demonstrated in Section~\ref{sec:results}).

Our clustering method (Section~\ref{sec:pairwise_clustering}) is based on the same idea, but it combines \emph{partial maps} instead of \emph{global maps}, therefore avoiding the mentioned problems of artifacts due to only partial consistency. On the technical side, the main challenge is that we cannot measure the distance between all pairs of candidate maps but only between actually overlapping ones. This is a problem for the original spectral clustering, which we substitute by an alternative technique geared towards sparse pairwise constraints \cite{zhang_gdl_eccv2012}.

\subsection{Previous Work on Local Isometry}

In previous work, there have been a number of attempts to find local isometric mappings, similar to our approach. However, these did not consider mappings between general surfaces but only local planar parametrizations.

Different techniques have been proposed to locally parameterize the intrinsic geometry of a surface to a plane using a local approximate isometry criterion.
Schmidt et al.~\cite{Schmidt2006} used exponential maps to transport a local coordinate system along a surface for the purpose of texture mapping. More recently, Schmidt~\cite{Schmidt2013} used transported exponential maps to produce a parameterization of a local surface patch to the plane that has low metric distortion. The local surface patch is provided through user interaction as an input stroke on the surface. 
Malvaer and Reimers~\cite{malvaer_reimers_2012} propose an alternative parameterization based on an extension of polar coordinates to surfaces. 
In our work, we cross-parameterize local surface patches from $\mathcal{S}$ to $\mathcal{T}$. This cross-parameterization is more challenging than a parameterization to the plane due to the arbitrary geometry of $\mathcal{T}$. Our cross-parameterization task is further complicated as in our application scenario, both $\mathcal{S}$ and $\mathcal{T}$ are scanned point clouds with missing surface information and scanner noise. Hence, the reviewed methods cannot be applied in a straight forward manner in our application.

\section{Local Metric Matching}
\label{sec:proposed}

We saw that for general surface matching using a global approximate isometry criterion for partial matching (Eq.~(\ref{eq:globalApproxIsoXu})) is difficult since the domain changes, and that using a local approximate isometry criterion (Eq.~(\ref{eq:localApproxIso})) is difficult since this results in a large search space.
In this section, we outline our new method:
starting from a point $\vecs \in \surfaceS$ and attached direction in $T_{\vecs} \surfaceS$ with known correspondence $f(\vecs) \in \surfaceT$ and corresponding attached direction in $T_{f(\vecs)} \surfaceT$,
we find the largest domain $\mathcal{U}$ such that Eq.~(\ref{eq:absoluteErrorPartial}) is satisfied. 

The key assumption of our approach is that $\surfaceS$ and $\surfaceT$ can be matched using a global approximate isometry in some partial region $\mathcal{U}$ containing $\vecs$, implying that Eq.~(\ref{eq:absoluteErrorPartial}) holds. Our goal is to find the largest region $\mathcal{U}$ for which this assumption is satisfied. This assumption holds in scenarios  where we know that $\surfaceS$ and $\surfaceT$ are actually related by a near-isometric map but the data does not comprise all of the original input and/or contains additional unrelated geometry or contacts. The most important practical example where this assumption holds is the acquisition of a surface that deforms near-isometrically but the scanning equipment introduces areas of missing data (shadowed to the scanner by other object parts) and cannot correctly resolve the topology in contact areas (such as the hand of a person being in contact with the body).

Our approach can be viewed as a hybrid approach between global and local approximate isometric matching. We use the assumption that $\surfaceS$ and $\surfaceT$ can be matched using a global approximate isometry in some partial region $\mathcal{U}$, and we compute $\mathcal{U}$ by growing a region 
%around $\vecs$ 
using a local approximate isometry criterion. This allows us to combine the advantage of the global methods of having a low-dimensional search space with the advantage of the local methods of being well suited to describe partial isometric matches.

Let $\Theta$ denote the parameter domain of all partial isometric matchings between $\surfaceS$ and $\surfaceT$. Our goal is to compute all partial isometries $\{f_{\theta},\mathcal{U}_{\theta}\}_{\theta \in \Theta}$ that map maximal subsets $\mathcal{U}$ of $\surfaceS$ to $\surfaceT$. The vector $\theta \in \Theta$ parametrizes the set of all such mappings.

In the following, we will discuss that:
\begin{itemize}
	\item Isometric deformations of $\surfaceS$ have three degrees of freedom.
	\item A partial isometry can be parametrized by $\Theta = \surfaceS \times SO(1) \times \surfaceT \times SO(1)$, where $SO(1)$ denotes the unit circle.
	\item There exists a global representation redundancy that identifies all choices of parameters $\vecs \in \mathcal{U} \subset \surfaceS$ and $\vecd_{\vecs} \in T_{\vecs}\surfaceS$. 
\end{itemize}

\subsection{Three Degrees of Freedom}

For a (sufficiently) smooth Riemannian manifold $\surfaceS$ fixing one point $\vecs$ on $\surfaceS$, a tangential direction $\vecd_{\vecs}$ in $T_{\vecs}\surfaceS$, and a surface normal $\vecn_{\vecs}$ at $\vecs$ suffices to specify an isometric mapping~\cite{Berger2002}. For details on the smoothness criteria, refer to~\cite{Palais57}. The following proof sketch aims to give the intuition behind this statement for 2-dimensional surfaces embedded in $\mathbb{R}^3$ by showing that we can transfer the local coordinate frame defined by $\vecd_{\vecs}$ and $\vecn_{\vecs}$ to any point on $\surfaceS$ in a canonical way. 

We start with two definitions. The \emph{injectivity radius $\rho_{\vecs}\surfaceS$} at $\vecs$ is the largest radius, such that for any point $\vecx$ on $\surfaceS$ with geodesic distance less than $\rho_{\vecs}\surfaceS$ from $\vecs$, there is a unique shortest geodesic path between $\vecs$ and $\vecx$. The \emph{injectivity radius of $\surfaceS$} is defined as $\rho\surfaceS = \inf_{\vecs \in \surfaceS}\left(\rho_{\vecs}\surfaceS\right)$. 

For closed surfaces, the injectivity radius can be bounded from below by the minimum of $\pi/\sqrt{\sup K}$, where $K$ is the Gaussian curvature, and half of the length of the smallest periodic geodesic~\cite[Thm. 10]{Berger2002}. It follows that $\rho\surfaceS > 0$ holds for closed Riemannian surfaces with finite Gaussian curvature. For general surfaces with boundary, the injectivity radius may become zero. Note that in this paper, we consider Riemannian surfaces with finite Gaussian curvature with boundaries. The boundaries are present because the closed Riemannian surface of interest is only partially observed by the acquisition device. In this special case, the injectivity radius is still positive.

Imagine that $\surfaceS$ is covered by overlapping regions of intrinsic radius less than $\rho\surfaceS$. These regions are all topologically equivalent to disks.
Consider a disk $D$ containing a point $\vecs$ as shown in Figure \ref{fig:ParallelTransport}.
In a small neighborhood of $\vecs$, $D$ is arbitrarily close to the tangent plane $T_{\vecs}\surfaceS$. Note that $\vecs$, $\vecd_{\vecs}$, and $\vecn_{\vecs}$ fix a local orthonormal coordinate frame in $\mathbb{R}^3$. This coordinate frame can be transported to a point $\vecx: \vecx \in D,\vecx \neq \vecs$ along the (unique) shortest geodesic path $P_{\vecs \vecx}$ from $\vecs$ to $\vecx$ in $D$ by parallel transport, which can be thought of as repeatedly projecting the direction $\vecd_{\vecs}$ to the tangent planes of consecutive points along $P_{\vecs \vecx}$ in infinitesimally small steps (for details on parallel transport see for example Berger~\cite[Chapter 3.1]{Berger2002}). Let $\vecd_{\vecx}$ denote the transported direction. The transferred direction lies in the tangent plane $T_{\vecx}\surfaceS$, and can again be used to fix a local orthonormal coordinate frame at $\vecx$. This transfer can be repeated by chaining together disks until every point $\vecy$ on $\surfaceS$ was assigned a fixed tangential direction $\vecd_{\vecy} \in T_{\vecy}\surfaceS$ by parallel transport along a path connecting $\vecs$ and $\vecy$ that consists of an arbitrary but fixed sequence of (unique) shortest geodesic paths within the chained disks. Note that by construction, we only use intrinsic information to propagate the direction $\vecd_{\vecs}$ to the entire surface. Hence, the resulting local coordinate frames are invariant with respect to isometric deformations of $\surfaceS$ when encoded relative to fixed local coordinate systems $\vecu(\vecy), \vecv(\vecy), \vecn(\vecy)$ on $\surfaceS$.

This implies that any isometric deformation of an \emph{oriented} surface $\surfaceS$ can be specified using three degrees of freedom: one point $\vecs$ on $\surfaceS$ (accounting for two degrees of freedom) and a direction in the tangent plane of $\vecs$.

\resetlength{\thispicwidth}
\addtolength{\thispicwidth}{\columnwidth}

\resetlength{\unitlen}
\addtolength{\unitlen}{\thispicwidth}

\begin{figure}%
\centering%
\pgfsetxvec{\pgfpoint{\unitlen}{0}}%
\pgfsetyvec{\pgfpoint{0}{\unitlen}}%
\begin{tikzpicture}%

\node (A) at (0,0) [inner sep=0pt, above right] {\includegraphics[trim=0 0 0 20cm, clip, width=\thispicwidth]{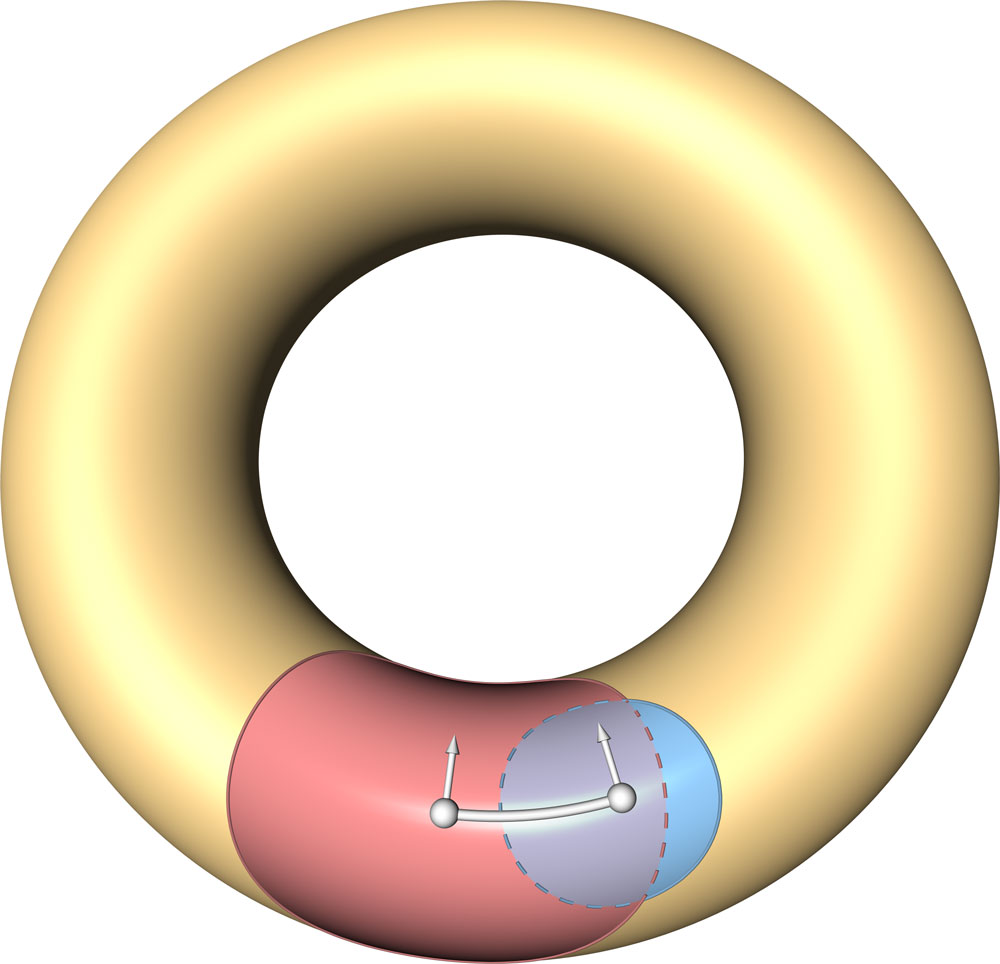}};

\begin{scope}[node distance=0, inner sep=0.01\unitlen, font=\large]
	\node (D) at (0.3, 0.2) [] {$D$};
	\node (s) at (0.44, 0.15) [below left] {$\vecs$};
	\node (x) at (0.63, 0.165) [below right] {$\vecx$};
	\node (p) at (0.525, 0.15) [below] {$P_{\vecs \vecx}$};
	\node (ds) at (0.45, 0.2) [above left] {$\vecd_{\vecs}$};
	\node (dx) at (0.61, 0.2) [above right] {$\vecd_{\vecx}$};
\end{scope}

\end{tikzpicture}%
\caption{Propagating a local coordinate system along $\surfaceS$.}%
\label{fig:ParallelTransport}%
\end{figure}

\subsection{Representation}
\label{sec:proposed_rep}

We can use the fact that any isometric deformation of an oriented surface can be specified using three degrees of freedom to derive a representation $\theta$ for intrinsic mappings. Specifically, identifying one corresponding point and one corresponding tangential direction completely determines an isometric mapping between two oriented surfaces. More formally, to define an isometric mapping between (subsets of) $\surfaceS$ and $\surfaceT$, it suffices to specify a point $\vecs \in \surfaceS$, its intrinsically corresponding point $\vect \in \surfaceT$, a tangential direction $\vecd_{\vecs}$ in $T_{\vecs}\surfaceS$, and its intrinsically corresponding tangential direction $\vecd_{\vect}$ in $T_{\vect}\surfaceT$.

Starting from this information, and assuming that $\surfaceS$ and $\surfaceT$ are isometric, we can propagate the correspondence information by mapping the metric structure of $\surfaceS$ onto $\surfaceT$ as follows. Starting from the corresponding points $\vecs$ and $\vect$ along with the corresponding directions $\vecd_{\vecs}$ and $\vecd_{\vect}$, we can propagate the correspondence information to a sufficiently small geodesic neighborhood $N_{\vecs}$ of $\vecs$ by simultaneously walking along corresponding geodesic paths starting at $\vecs$ and $\vect$, respectively, and by matching points that are reached at the same time. Here, $N_{\vecs}$ is sufficiently small if its geodesic radius is below $\rho\surfaceS$. Once the correspondence information is computed for $N_{\vecs}$, we continue propagating the correspondence information from a point close to the boundary of $N_{\vecs}$ to its geodesic neighborhood, and iterate until every point on $\surfaceS$ has a correspondence.

The assumption that $\surfaceS$ and $\surfaceT$ are (near-) isometric can also be used to detect the boundary of the largest region $\mathcal{U} \subseteq \surfaceS$ containing $\vecs$ for which the mapping is near-isometric. The reason is that the propagation algorithm allows us to measure the difference in intrinsic geometry in newly mapped parts of $\surfaceS$ and $\surfaceT$ directly. Hence, we can stop the region growing algorithm if a newly added correspondence would induce a stretching larger than $\nu$.

\begin{figure}%
\centering%
\subfigure[Oriented point match.]%
{\label{fig:growinga}\includegraphics[width=\ltwopicwidth]{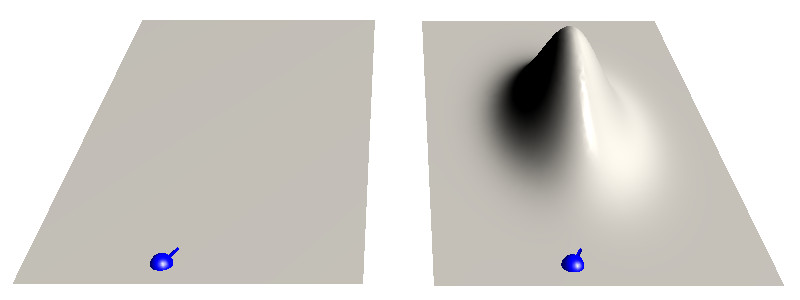}}%
\goodgap%
\subfigure[Some growing.]%
{\label{fig:growingb}\includegraphics[width=\ltwopicwidth]{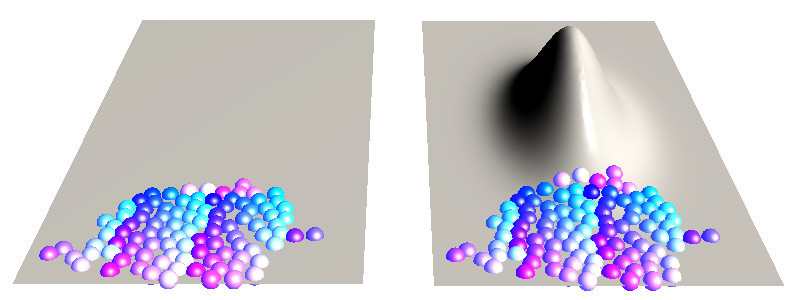}}%
\\
\subfigure[More growing.]%
{\label{fig:growingc}\includegraphics[width=\ltwopicwidth]{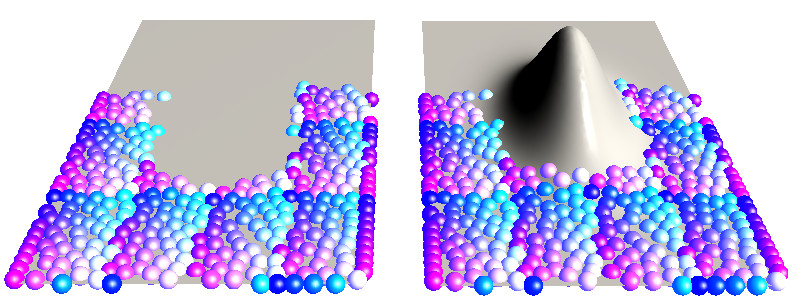}}%
\goodgap%
\subfigure[Final mapping.]%
{\label{fig:growingd}\includegraphics[width=\ltwopicwidth]{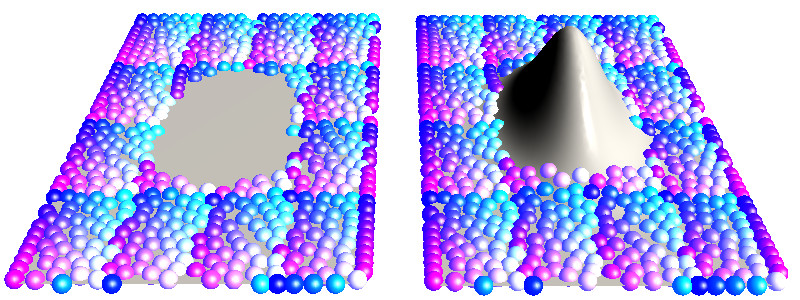}}%
\caption{Illustration of the near-isometric region growing process. Corresponding points share the same color.}
\label{fig:growing}
\end{figure}

Figure~\ref{fig:growing} illustrates the near-isometric region growing process. The plane on the left ($\surfaceS$) and the plane with a hill on the right ($\surfaceT$) are isometric except for the hill. Beginning with a point and direction match in Figure~\ref{fig:growinga}, the isometric region grows outward in all directions where the isometry condition is locally satisfied. This way, the largest near-isometric partial match is identified, as shown in Figure~\ref{fig:growingd}. 
Note that this region can have a complex topology and geometry.

\subsection{Representation Redundancy}

The representation discussed above contains redundant information: Infinitely many $\theta$ may represent the same near-isometric mapping.

The first degree of redundancy is the choice of the direction $\vecd_{\vecs}$ in the tangent plane $T_{\vecs}\surfaceS$. Changing this direction merely rotates the field of directions $\vecd_{\vecx}$ in $T_{\vecx}\surfaceS$ (and similarly, the field of corresponding directions $\vecd_{\vecy}$ in $T_{\vecy}\surfaceT$). Hence, the choice of this direction has no influence on the final result. To remove this redundancy, we start from an arbitrary but fixed direction $\vecd_{\vecs}$ and precompute and fix $\vecd_{\vecx}$ for all $\vecx$ on $\surfaceS$.

The second redundancy is the choice of the start point $\vecs$. Let $\mathcal{U} \subseteq \surfaceS$ be the maximal set within which an isometry $f: \mathcal{U} \rightarrow \surfaceT$ can be constructed.
We can replace $\vecs$ in $\theta$ by any other point $\vecs' \in \mathcal{U}$.
See Fig.~\ref{fig:Redundancy}.
This requires an update of the remaining parameters. The direction $\vecd_{\vecs'}$ is set to the precomputed value (see above), and the remaining parameters can be updated using the computed mapping $f$. More specifically, $\vect' = f(\vecs')$ and the direction of $\vecd_{\vect'}$ is set to the direction of $f(\vecs' + \epsilon \vecd_{\vecs'}) - f(\vecs')$, where $\epsilon$ is set small enough that $\vecs' + \epsilon \vecd_{\vecs'}$ is in $\mathcal{U}$. 
Thus, in the case of a global isometry, or when we know beforehand the isometric region $\mathcal{U}$, the mapping $f$ has three degrees of freedom. In the general partial isometry case, however, where $\mathcal{U}$ is unknown, the starting point $\vecs \in \surfaceS$ is not fully redundant; it still selects the equivalence class that represents a certain partial map; computationally, this is the patch to be matched by the propagation algorithm. This does not increase the complexity strongly as we just have to restart the matching in case multiple partial matches exist. Ideally, we would sample one starting point per partial isometric region, which in practice will be far fewer than there are samples on $\surfaceS$. While we do not determine this lower-bound beforehand, we maintain low complexity by using features to identify potential starting points, and marking a starting point $\vecs$ as redundant if another starting point $\vecs'$ produces an isometric region $\mathcal{U}$ containing $\vecs$. In case of a mismatch, it can be discovered quickly if the target area does not match, which will become evident within constant time.

\resetlength{\thispicwidth}
\addtolength{\thispicwidth}{\columnwidth}

\resetlength{\unitlen}
\addtolength{\unitlen}{\thispicwidth}

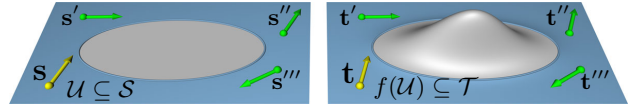
\begin{figure}%
\centering%
\pgfsetxvec{\pgfpoint{\unitlen}{0}}%
\pgfsetyvec{\pgfpoint{0}{\unitlen}}%
\begin{tikzpicture}%

\node (A) at (0,0) [inner sep=0pt, above right] {\includegraphics[width=\thispicwidth]{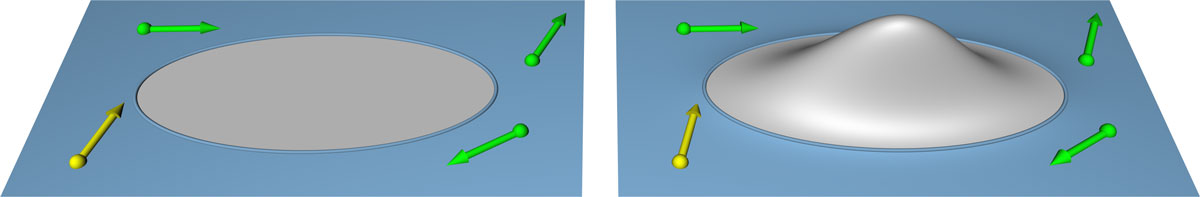}};

\begin{scope}[node distance=0, inner sep=0.01\unitlen, font=\large]

	\node (s) at (0.05, 0.05) [] {$\vecs$};
	\node (t) at (0.55, 0.05) [] {$\vect$};
\end{scope}

\begin{scope}[node distance=0, inner sep=0.0\unitlen, font=\small]
	\node (U) at (0.15, 0.005) [above] {$\mathcal{U} \subseteq \surfaceS$};
	\node (T) at (0.675, 0.005) [above] {$f(\mathcal{U}) \subseteq \surfaceT$};

	\node at (0.1, 0.145) [] {$\vecs'$};
	\node at (0.55, 0.145) [] {$\vect'$};
	
	\node at (0.46, 0.145) [left] {$\vecs''$};
	\node at (0.91, 0.145) [left] {$\vect''$};

	\node at (0.45, 0.04) [] {$\vecs'''$};
	\node at (0.95, 0.04) [] {$\vect'''$};
\end{scope}

\end{tikzpicture}%
\caption{For a given set $\mathcal{U}$ and a corresponding isometry (shown in blue), the choice of the starting point $\vecs$ is arbitrary.}%
\label{fig:Redundancy}%
\end{figure}

In summary, all mappings represented by $\vecs', \vecd_{\vecs'}, \vect', \vecd_{\vect'}$ form an equivalence class in the parameter space $\Theta$. Since we can remove one degree of redundancy by fixing the tangential directions on $\surfaceS$, for each mapping $f$ we have two degrees of freedom that vary among equivalent maps (the choice of the start point on $\surfaceS$), which along with the manifold structure of $\mathcal{U}$ implies that each equivalence class forms a $2$-manifold in $\Theta$, which can be computed directly using the propagation algorithm introduced in Section~\ref{sec:proposed_rep}.

In practice, we can take advantage of this representational redundancy as follows. Since $\surfaceS$ and $\surfaceT$ are discretized and corrupted by noise, the error of the correspondence information computed using the propagation algorithm increases with increasing distance from the start point of the propagation.
Hence, it is possible that the propagation stops prematurely due to discretization artifacts and the influence of noise, thereby identifying a region $\mathcal{U}$ that is smaller than the correct solution.
Thanks to the redundancy in the representation, we can start the propagation algorithm from multiple oriented point pairs, identify a set of equivalent mapping functions $f_i$, and them into a single mapping function $f$ that covers a larger area of $\surfaceS$ and is less influenced by noise than the individual $f_i$.
The following section discusses a direct implementation of this theoretically motivated method, which we will use to compute correspondences of noisy scanner data.

\section{Pairwise Intrinsic Matching}
\label{sec:pairwise}

\begin{figure*}[htb]
\centering
\includegraphics[width=0.9\textwidth]{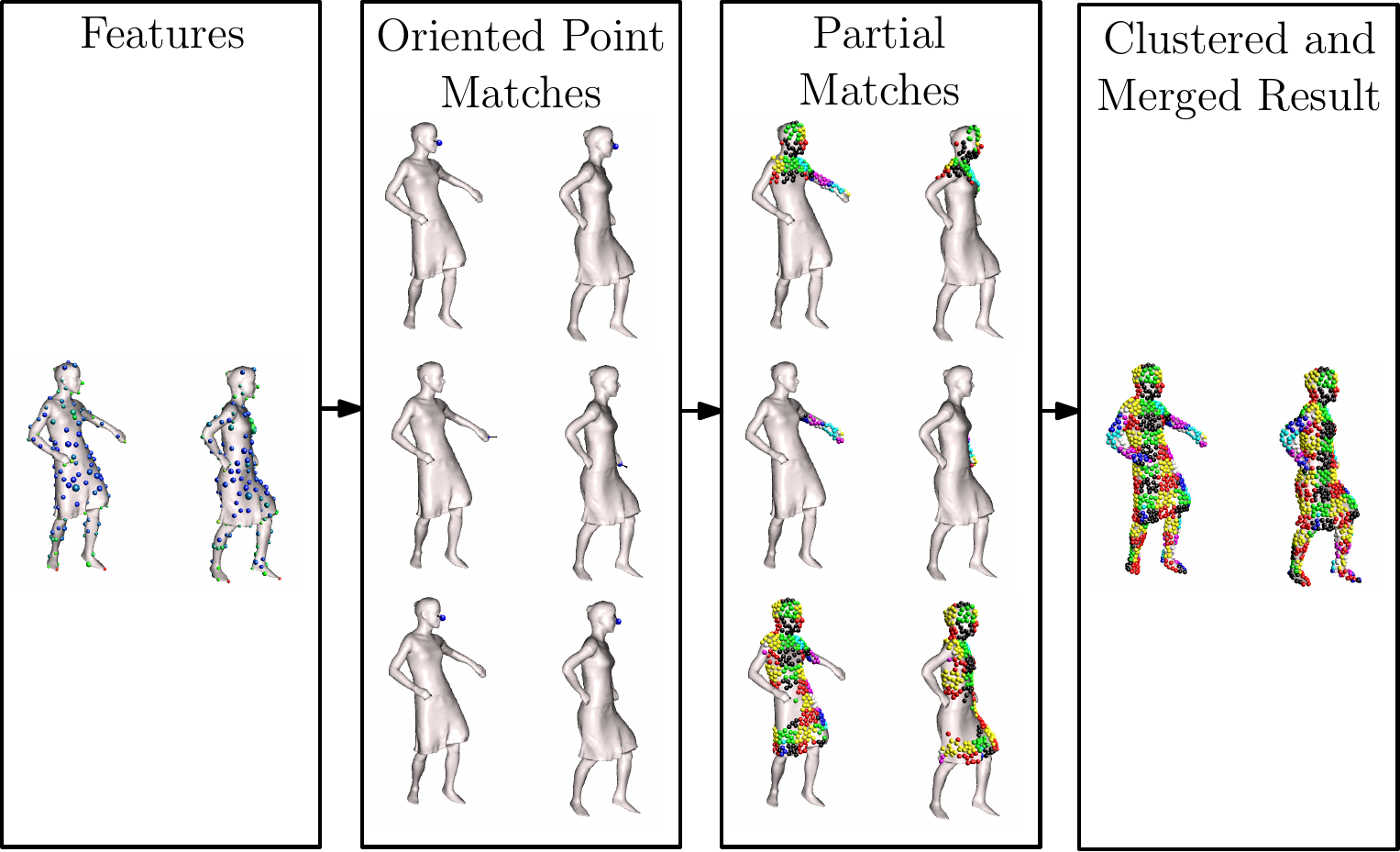}
\caption{Overview of the pairwise matching pipeline. First, features are detected, second, oriented point matches are computed, third, starting from the oriented point matches, partial isometric matches are found, and finally, different partial matches are clustered and merged.}
\label{fig:overview_matching}
\end{figure*}

From the preceding analysis, we derive an algorithm for computing partial near-isometries between two surfaces $\surfaceS$ and $\surfaceT$. We start our discussion by outlining how surfaces are represented and how basic steps of the algorithm are implemented (Section~\ref{sec:implement_basic}). Our direct, non-optimized implementation is based on enumerating the non-equivalent choices by selecting different oriented point matches (Section \ref{sec:pairwise_pointmatches}), growing the isometric region locally from there until no more points locally satisfy $\nu$ (Section \ref{sec:pairwise_growing}), and finally clustering the partial maps into equivalence classes (Section \ref{sec:pairwise_clustering}). Figure~\ref{fig:overview_matching} gives a graphical overview of our matching pipeline.

\subsection{Surface Representation}
\label{sec:implement_basic}

In our implementation, the surfaces $\surfaceS$ and $\surfaceT$ are either represented as oriented point clouds or meshes. While most parts of the algorithm can be extended to discretized surfaces in a straight forward manner, for some parts of the algorithm we require a continuous surface representation. We obtain a continuous surface representation as implicit moving-least-squares (MLS) manifolds using the robust method of \"{O}ztireli et al.~\cite{oeztireli_eg2009}. Using this method, a discretized surface $\surfaceS$ is represented continuously as the zero level-set of an implicit function derived from the oriented vertices of $\surfaceS$. 

Using a MLS representation allows a point to be projected to a continuous representation of $\surfaceS$ in the case where $\surfaceS$ is given as oriented point cloud or mesh. In the following, whenever we refer to projecting a point $\vecx$ onto a discretized surface $\surfaceS$, this projection is implemented as projecting $\vecx$ to the MLS representation derived from $\surfaceS$.

One basic operation needed by our algorithm is the computation of geodesic distances and paths on $\surfaceS$. In some parts of the algorithms, rough estimates of geodesic distances and paths suffice, and these are computed using Dijkstra's algorithm. In other parts of the algorithm, it is crucial to have accurate estimates of geodesic distances and paths. In these cases, we initialize a geodesic path $P$ to the Dijkstra path and refine $P$ by minimizing the length of $P$ using the constraint that $P$ must not leave $\surfaceS$. This optimization is carried out iteratively using a conjugate gradient method. After each step, $P$ is projected back to $\surfaceS$. In the following, we refer to these refined geodesic distances and paths as \emph{smoothed geodesic distances and paths}.

A core part of our approach is to compare distances measure on the target surface to distances measured on the source for corresponding points. Comparing geodesics between all pairs of correspondences quickly becomes prohibitively expensive and would limit the practical applicability of our algorithm. To remedy this, we construct a topology hierarchy on $\surfaceS$ similar to \cite{memoli05} as follows. We define level $0$ of the hierarchy to be the original set of vertices and their connectivity--either the original triangle mesh or the $k$-nn graph for a point cloud. (In all our experiments involving point clouds, we set $k=8$.) The sample spacing $\epsilon_0$ is defined as the average edge length. Level $1$ of the hierarchy is constructed by selecting an evenly spaced subset of the level $0$ vertices and connecting vertices in a topology preserving way. Subsequent levels $j+1$ are constructed in the same way as a subset of level $j$. At each level the subset for the next level is determined by doubling the desired sample spacing, $\epsilon_{j+1}=2\epsilon_{j}$. At a coarser level of this hierarchy fewer vertices are connected to any others, but at a greater distance to each other. In our algorithm, we only consider geodesic distances between vertices that share an edge in at least one level of the hierarchy, which results in a sparse set of distances to be optimized, even for a dense set of correspondences.

This hierarchical structure ensures that locality is respected, as only geodesics between points that are neighbors in some level of the hierarchy are considered. Having a structure that respects locality is crucial to allow for partial matching. The number of levels in the hierarchy determines the trade-off between local and global isometry constraints. We use $5$ levels in all experiments in this paper.

Note that during the execution of the matching algorithm, the vertices on $\surfaceS$ are fixed, as we aim to find a correspondence for every vertex of $\surfaceS$ on $\surfaceT$ (if such a correspondence exists). Hence, we can precompute all geodesic distances on $\surfaceS$ for vertices connected in some level of the hierarchy. This way, only distances on $\surfaceT$ need to be updated during the metric optimization.

\subsection{Finding Oriented Point Matches}
\label{sec:pairwise_pointmatches}

In principle, by exhaustively trying all possible starting points and directions, our algorithm will recover all partial isometries. However, since this is infeasible, we reduce the search space using sparse feature matches. This section outlines an algorithm to find features matches. However, note that this part is not a novel contribution of this paper and only given for completeness; in principle, any feature matches can be used to reduce the search space, as demonstrated in Section \ref{sec:results}, where we show a result using image-based feature matches from a multi-view capture setup. We denote the feature descriptor at a point $\vecs\in\surfaceS$ with a vector $\vecD_{\surfaceS}(\vecs)$, and similarly for points on $\surfaceT$.

We identify features on $\surfaceS$ and $\surfaceT$ using the geodesic fingerprint descriptor $\vecD_{\surfaceS}(\vecs)$~\cite{sun01}, which compares how an isocontour of the geodesic distance from a point $\vecs \in \surfaceS$ deviates in length from the isocontour of the same 2D Euclidean distance. We use $10$ isocontours in all experiments, for geodesic radii between $r_{\min}$ and $r_{\max}$. The values of $r_{\min}$ and $r_{\max}$ need to be varied slightly depending on the amount of noise in the input data, and are discussed in Section \ref{sec:results}. We define the \emph{distinctiveness $F_{\surfaceS}(\vecs)$} of a point $\vecs \in \surfaceS$ as the sum of $L_1$ distances to the rest of the vertices of $\surfaceS$ in descriptor space. Features are points that locally maximize distinctiveness. The left-most box in Figure \ref{fig:overview_matching} shows features color-coded by distinctiveness (red for most distinct, blue for least distinct).

To find feature matches, we begin by computing the Cartesian product of $L_2$ descriptor distances between features on $\surfaceS$ and $\surfaceT$. The vast majority of these are not correct mappings between the surfaces. We filter these potential feature matches using both the $L_2$ distances between descriptors and the distinctiveness of the features. More precisely, for each feature $\vecs$ on $\surfaceS$, we only consider the $K$ features of $\surfaceT$ that have the closest descriptor matches to $\vecs$. (We use $K=10$ in all our experiments.) The \emph{initial dissimilarity $\delta_{\vecs,\vect}^{(init)}$} between two features $\vecs \in \surfaceS$ and $\vect \in \surfaceT$ is defined as 
{\scriptsize
\begin{equation}
\delta_{\vecs,\vect}^{(init)} = -\log\left(F_{\surfaceS}(\vecs)\right) + \omega_{\vecD}\left\|\vecD_{\surfaceS}(\vecs) - \vecD_{\surfaceT}(\vect)\right\|_2^2,
\end{equation}
}
where $\omega_{\vecD}$ is a weight. (We use $\omega_{\vecD}=400$ in all our experiments.) This dissimilarity is minimized for features with high distinctiveness that are similar in descriptor space.

This procedure often produces a good set of sorted feature matches on clean data. For noisy data from real scanners, however, the descriptors will be less discriminative, and considering spatial relations between features in addition to descriptors and distinctiveness leads to more reliable feature matches.

To do this, we iteratively build (possibly overlapping) clusters $C_i$ of consistent feature matches. In each iteration, the feature match $(\vecs, \vect)$ with the next lowest $\delta_{\vecs,\vect}^{(init)}$ is chosen as a starting point for $C_i$. The cluster $C_i$ is built by repeatedly adding the close-by feature match that has the lowest dissimilarity to $C_i$. More precisely, let $C_i = \{(\vecs_j,\vect_j)\}$. In the next step, all features $\vecs' \not\in C_i$ that are neighbors of $\vecs_j$ in the topology hierarchy are considered, along with their $K$ potential matches $\vect' \in \surfaceT$. The dissimilarity $\delta_{C_i,(\vecs',\vect')}$ between a feature match $(\vecs',\vect')$ and cluster $C_i$ is defined as 
{\scriptsize
\begin{equation}
\delta_{C_i,(\vecs',\vect')} = \delta_{\vecs,\vect}^{(init)} + \omega_{C} \sum_{(\vecs_j,\vect_j)} \left|\dist_{\surfaceS}(\vecs_j, \vecs') - \dist_{\surfaceT}(\vect_j, \vect')\right|^2,
\end{equation}
}
where $\omega_{C}$ is a weight. (In all our experiments, we set $\omega_{C}$ to one over $8\times$ the square of the average edge length on $\surfaceS$.) We repeatedly add the match $(\vecs',\vect')$ with smallest dissimilarity $\delta_{C_i,(\vecs',\vect')}$ to $C_i$ as long as $\delta_{C_i,(\vecs',\vect')}$ is below a threshold. (We use threshold $11.5\approx-\log 10^{-5}$ in all our experiments.) We stop adding new clusters once the sum of the cardinalities of the clusters $C_i$ exceeds the initial number of feature matches.

Note that the above clustering scheme is equivalent to modeling both the descriptor distances and the summed stretches of geodesic distances of a feature match to a cluster as normally distributed. Hence, the above stopping criteria correspond to stopping the clustering once the joint probability of a match belonging to a cluster becomes small.

After the clustering, we place the feature matches $(\vecs, \vect)$ in a min-priority queue, so that we start isometric region growing first from the matches we expect to be most reliable. A feature match $(\vecs, \vect)$ is assigned priority $\infty$ if $(\vecs, \vect)$ is not part of any cluster $C_i$ and priority $\min_{C_i: (\vecs, \vect)\in C_i} \delta_{C_i,(\vecs',\vect')}$ otherwise. We repeatedly take the minimum element from queue and use it to generate partial isometric mappings using the region growing of Section \ref{sec:pairwise_growing}. 

However, so far we have only established a positional correspondence between features, and we need a directional correspondence as well to fix the partial isometry we wish to grow. To establish direction, we build a min-priority queue as outlined above, but only with the subset of features in the neighborhood of the current positional match in the topology hierarchy. Matches of neighboring features allow us to find corresponding tangent plane directions from corresponding smoothed geodesic paths between feature points.

Once a partial isometry has been computed, we increase the priority of feature matches that are redundant given the already computed partial match. If the same source and target points are already matched, by our model, the result of running the growing again from those same points will be equivalent.

To keep the run-time of our method bounded, we stop after a fixed number of oriented point matches have been tried. (We use $200$ in all our experiments.) 
A more general stopping criterion could be devised based on determining the maximal coverage of $\surfaceS$ and $\surfaceT$ subject to a consistent mapping. The second box from the left of Figure \ref{fig:overview_matching} shows three oriented feature matches found using this method.

\subsection{Isometric Region Growing}
\label{sec:pairwise_growing}

Starting from an oriented point match $\vecs,\vecd_{\vecs},\vect,\vecd_{\vect}$, we grow the region $\mathcal{U}$ by adding matches incrementally in the local neighborhood of the boundary of $\mathcal{U}$.
For a new point $\vecu$ near the boundary of $\mathcal{U}$ such that $\vecu\notin\mathcal{U}$, let $N_1(\vecu)$ denote the neighborhood of $\vecu$ in level $1$ of our topology hierarchy. We use parallel transport along corresponding directions in $\surfaceS$ and $\surfaceT$ emanating from an oriented point match $\vecs',\vecd_{\vecs'},\vect',\vecd_{\vect'}$, where $\vecs'\in\mathcal{U}\cap N_1(\vecu)$ is in a local neighborhood of $\vecu$. 
We know the full path between $\vecs'$ and $\vecu$ on $\surfaceS$, and from $\vecd_{\vecs'}$ and $\vecd_{\vect'}$ we know the corresponding start direction on $\surfaceT$. 
Hence, we can transport the start direction along corresponding paths on $\surfaceS$ and $\surfaceT$ until we have traveled $\dist_{\surfaceS}(\vecs',\vecu)$. 
In practice, we implement parallel transport by moving in small steps in the tangent plane, and by reprojecting the resulting point to the surface and updating the tangent to the path. We use smoothed geodesic paths to transport the position of the new match on $\surfaceT$ to reduce discretization errors for differently sampled surfaces. For robustness, we use all available oriented point matches in $N_1(\vecu)$ and take the Riemannian center of mass~\cite{rustamov_sgp2010} of the transported points, where all transported points have equal mass. 

The newly matched source point $\vecu$ is added to $\mathcal{U}$ only if it locally respects the stretch factor $\nu$ as given in Eq.~(\ref{eq:absoluteErrorPartial}). This is necessary for two reasons. First, subsequent matches will be estimated based on the assumption that the existing matches are near-isometric. Second, as discussed next, we use a nonlinear least-squares optimization to refine the positions of the matched points on $\surfaceT$, which effectively distributes the error evenly over the matched region. To avoid introducing errors, it is therefore important to verify that each newly added point match locally respects the stretch factor $\nu$.

To reduce the effect of quantization errors and noise, we optimize the metric matching of $\mathcal{U}$ using a non-linear optimization technique~\cite{Bronstein2006} every time the area of $\mathcal{U}$ has doubled. This means more frequent optimizations at the start of the growing process. This optimization reduces the amount of drift as it re-aligns the matched regions while taking into account long geodesics, between neighbors in the top level of our topology hierarchy, in $\mathcal{U}$. For increased efficiency, we only consider edges in the topology hierarchy of $\surfaceS$ (explained in Section~\ref{sec:implement_basic}) during the optimization.

An important difference to the global approach used by Bronstein et al.~\cite{Bronstein2006} is that we optimize the metric only using geodesic paths which are entirely \emph{within} $\mathcal{U}$ and $f(\mathcal{U})$. This models the isometry criterion in partial regions and is crucial to handling topology changes and missing data. Note that such defects may cause large differences between $\dist_{\surfaceS}(\vecx,\vecy)$ and $\dist_{\mathcal{U}}(\vecx,\vecy)$, as well as $\dist_{\surfaceT}(f(\vecx),f(\vecy))$ and $\dist_{f(\mathcal{U})}(f(\vecx),f(\vecy))$, respectively. The second box from the right in Figure \ref{fig:overview_matching} shows the first three partial isometries found by growing isometrically from oriented feature matches in our priority queue.

The proper value for the stretching threshold $\nu$ depends on a number of factors: material properties, the resolution at which the surface is sampled (as it affects the accuracy of the Dijkstra paths), and the noise of the acquisition process. In our experiments, we do not consider material properties, and we assume that the acquisition noise has an equal influence on both source and target. Hence, we set $\nu$ to $0.5\epsilon_0$ to account for quantization effects in the computation of geodesics. 

\subsection{Combining Equivalent Partial Maps}
\label{sec:pairwise_clustering}

It remains to identify and merge a set of partial mappings that represent the same mapping function $f$. If the surfaces were related by exact isometries and noise was negligible, the following step would not be required. However, for real-world data, the identification of functions that are approximately equivalent improves the results substantially.

The problem at this point similar to the blending problem by Kim et al.~\cite{Kim2011}. Recall that their approach uses a spectral method to find blending weights for different maps. This is a good approach in their case as blending weights are given as the solution to a quadratic energy function. Note that since in our model, equivalent mappings form a 2-manifold in parameter space $\Theta$, and the parameter space $\Theta$ is non-linear, it is not appropriate to use a spectral approach.

However, we can take advantage of the property that equivalent mappings form a 2-manifold in $\Theta$. We employ the agglomerative clustering algorithm of Zhang et al.~\cite{zhang_gdl_eccv2012} for discovering manifold structures in high-dimensional data based on the in-degree and out-degree of the nearest-neighbor graph of points in high-dimensional space. In our case, we consider the nearest neighbor graph of partial isometric mappings, where the dissimilarity between these points in $\Theta$ is measured as follows. 

We compare different maps $f_i$ and $f_j$ using a dissimilarity measure based on their domains $\mathcal{U}_i$ and $\mathcal{U}_j$:
{\scriptsize
\begin{eqnarray}\nonumber
d_{\Theta}(f_i,f_j) & = & W_1 \int_{\mathcal{U}_{ij}} \dist_{\surfaceT}(f_i(\vecx), f_j(\vecx)) d\vecx \\ 
 & + & W_2 \int_{f_i(\mathcal{U}_{ij})\cap f_j(\mathcal{U}_{ij})} \dist_{\surfaceS}(f_i^{-1}(\vecy), f_j^{-1}(\vecy)) d\vecy
\end{eqnarray}
}
where $\mathcal{U}_{ij}=\mathcal{U}_i\cap\mathcal{U}_j$, $W_1=\frac{1}{A(\mathcal{U}_{ij})}$, $W_2=\frac{1}{A(f_i(\mathcal{U}_{ij})\cap f_j(\mathcal{U}_{ij}))}$, and $A(\cdot)$ denotes the surface area. In practice, we compute a discrete version of this dissimilarity by replacing integrals over regions by sums over vertices in the region. We cluster different mappings together until the maximum affinity between any two clusters is greater than a threshold $\rho$. Affinity is computed from the weighted graph degree between mappings, where the weights have a double-exponential fall-off as $d_{\Theta}$ increases. See Zhang et al.~\cite{zhang_gdl_eccv2012} for details. While the direct relation of $\rho$ to the allowed stretch is not easily defined, it should be lower when we want to allow for greater stretching between partial maps. We only have to adjust this value in a few cases in our experiments, as discussed in Section \ref{sec:results}. Following the clustering, we select for our final mapping the cluster with the highest intra-cluster affinity, or connectivity, as proposed by Zhang et al.~\cite{zhang_gdl_eccv2012}. This most often correlates with the cluster that covers the largest portion of the surface.

We merge clustered maps by computing a weighted geodesic average on $\surfaceT$ for each source point in the union of the mapped regions as
{%\scriptsize
\begin{equation}
f(\vecx) = \arg\min_{y\in\surfaceT} \sum_i w_i(\vecx) \dist_{\surfaceT}(\vecy, f_i(\vecx)),
\end{equation}
}
where $f_i$ are the clustered partial isometries, which we want to merge into $f$. The weights are computed as an exponential distribution in the geodesic distance from the starting point match $\vecs_i$ as
$w_i(\vecx) = \exp(-\lambda_{ds} \dist_{\mathcal{U}_i}(\vecx, \vecs_i))$,
reflecting that we expect errors to accumulate as the growing proceeds because of discretization artifacts, data noise, and deviations from isometry. We set $\lambda_{ds}=1/(5\epsilon_1)$, where $\epsilon_1$ is the sample spacing of the level $1$ of the topology hierarchy described in Section \ref{sec:implement_basic}. Note that this is equivalent to finding the Riemannian center of mass~\cite{rustamov_sgp2010} of the estimates $f_i(\vecx)$. The right-most box of Figure \ref{fig:overview_matching} shows the final clustered and merged result.

\section{Experiments}
\label{sec:results}

\resetlength{\thispicheight}
\addtolength{\thispicheight}{4cm}
\resetlength{\thispicwidth}
\addtolength{\thispicwidth}{2.05cm}

To validate our theoretical analysis, we evaluate a direct implementation of our algorithm and compare our results to state of the art approaches. In the following, we demonstrate that our method achieves results that are either comparable or superior to the state of the art, which demonstrates that our algorithm not only has theoretical advantages, but is applicable in practice as well.

\subsection{Implementation Details}

We implemented the algorithm described in Section~\ref{sec:pairwise} in C++, and conducted our evaluations on a standard laptop PC. During the evaluation, all but two types of parameters are fixed. The first type of parameters that is varied is $\{r_{min},r_{max}\}$, which controls the size of the neighborhood used to compute surface descriptors. If the radii are set higher, then the method is more robust with respect to noise at the cost of potentially missing features of small scale. In our experiments, we only use two settings for these parameters. The first setting, $r_{min}=0.9R$ and $r_{max}=1.7R$, is for relatively clean data, and the second setting, $r_{min}=1.5R$ and $r_{max}=3.4R$, is for noisy data. Here, $R$ is set to $5\%$ of the diameter of $\surfaceS$. The second parameter that is varied is the threshold $\rho$ used to control the clustering. Lower values of $\rho$ allow for less isometric patches to be clustered together. 
%Again, we only use two settings for this parameter. We set $\rho=1$ for data that is fairly isometric and $\rho=0.1$ for data that exhibits a significant amount of non-isometric stretching.
We vary $\rho\in\{0,1,1.9\}$ in our experiments.

For the examples discussed below, our algorithm takes between 30 minutes and 8 hours to compute the final result. To give an idea of the distribution of the time, we discuss the running time for one pair of models (the space-carved samba models shown in Fig.~\ref{fig_samba_contacts}) in more detail. For this pair, finding oriented feature matches takes about 2 minutes, growing partial mappings takes about 1.5 hours, clustering the mappings takes about 9 minutes, and merging the patches takes about 44 minutes. Hence, the total time to compute the results is about 2.4 hours. Note however, that the running time of our method depends significantly on the distinctiveness of the intrinsic geometry of the surfaces, relative to the noise level. For example, the template-fitted samba models shown in Fig.~\ref{fig_samba_clean} take about 1.5 hours in total, despite having more than twice as many vertices as the space-carved versions, because the approximate isometry criterion is more discriminative (geodesic distances are less perturbed by surface noise).

\subsection{Comparison to State of the Art}

We compare the performance of our algorithm against four existing methods, namely heat kernel maps (HKM)~\cite{Ovsjanikov2010}, blended intrinsic maps (BIM)~\cite{Kim2011}, the method of Sharma et al.~\cite{Sharma2011}, and the method of Tevs et al.~\cite{TevsAnimRec2012}. We select these methods for comparison because they represent the state of the art for matching between surfaces. More specifically, HKM is derived from the theoretical complexity of isometric mappings, BIM can match surfaces that exhibit local deviations from isometry, and the methods of Sharma et al. and Tevs et al. are heuristics that have been specifically designed for matching partial data with topological noise. For HKM we use our own implementation, which uses two feature correspondences to initialize the mapping, for BIM we use the authors' implementation, for comparisons with Sharma et al., we run our code on their data, and for the method of Tevs et al., the authors were kind enough to run their algorithm on our data.

We show comparative evaluations on a variety of types of data. The first type is a synthetic data set that helps in understanding the major difference between our approach and HKM, and illustrates the partial isometric matching model. The second type is data acquired using either a laser scanner or an image-based reconstruction system that was processed by fitting a template to the data. In this case, we treat the result of the template fitting as ground truth. The third, and most challenging, type is unprocessed real-world data acquired using different acquisition systems.

To compare our approach to previous methods, we use two evaluation methodologies. In cases where ground truth is available, we compare quantitatively by evaluating the accuracy of different results with respect to the ground truth. For data that has no ground truth, we rely on visual evaluation. Our visualization scheme is as follows. A texture on $\surfaceS$ is mapped to corresponding points on $\surfaceT$, and regions of $\surfaceT$ that have no correspondence in $\surfaceS$ are colored red. As a texture we combine constant coloring of semantically distinct parts (where applicable) with a checkerboard pattern. This type of texture simultaneously shows both global semantic accuracy and fine-scale distortion.

\subsection{Synthetic Data}

\begin{figure}%
\includegraphics[width=\columnwidth]{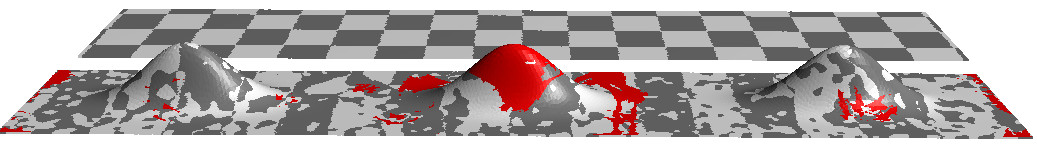}%
\\%
{\small\sffamily\makebox[\columnwidth]{Heat kernel maps \cite{Ovsjanikov2010}}}%
\\[\baselineskip]%New Line
\includegraphics[width=\columnwidth]{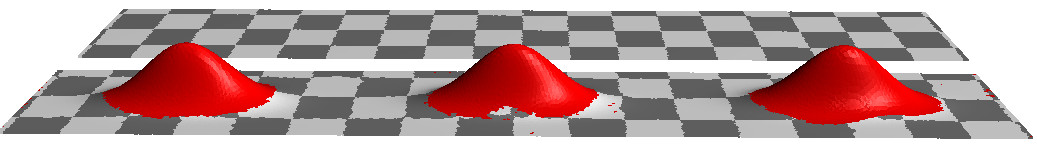}%
\\%
{\small\sffamily\makebox[\columnwidth]{This paper}}%
\caption{Results of our method and HKM on globally non-isometric data. Red indicates unmatched area.}
\label{fig_plane3peaks}
\end{figure}

We start by comparing our algorithm against HKM using the synthetic example shown in Fig.~\ref{fig_plane3peaks}, where we map from a plane to a plane with three peaks. This experiment illustrates the difference between a global isometric matching model and partial one: the two surfaces are globally non-isometric, but the planar part is isometric. In this example, we use two ground truth correspondences to initialize both algorithms to remove differences due to different feature matching approaches. Since HKM aims to map the shapes using a global isometry, the result maps planar parts to the peaks, while our method successfully detects the largest part of the surfaces that can be isometrically mapped.

\subsection{Template-Fitted Scan Data}

Next, we consider acquisitions of real-world data that was processed by fitting a template shape to the raw data. For all experiments in this section, we use $r_{min}=0.9R$ and $r_{max}=1.7R$.

We first use two frames of the samba sequence by Vlasic et al.~\cite{vlasic2008}. These frames are locally very close to isometric, but globally have high non-isometric distortion due to the dress being connected to the legs. We therefore set $\rho=0$ in the clustering step. Vlasic et al. provide a processed version of the frames, where a template was fitted to the data. This processing ensures that the models have the same topology, and the processed data can be used as ground truth correspondence. 

\begin{figure*}%
\includegraphics[height=\thispicheight]{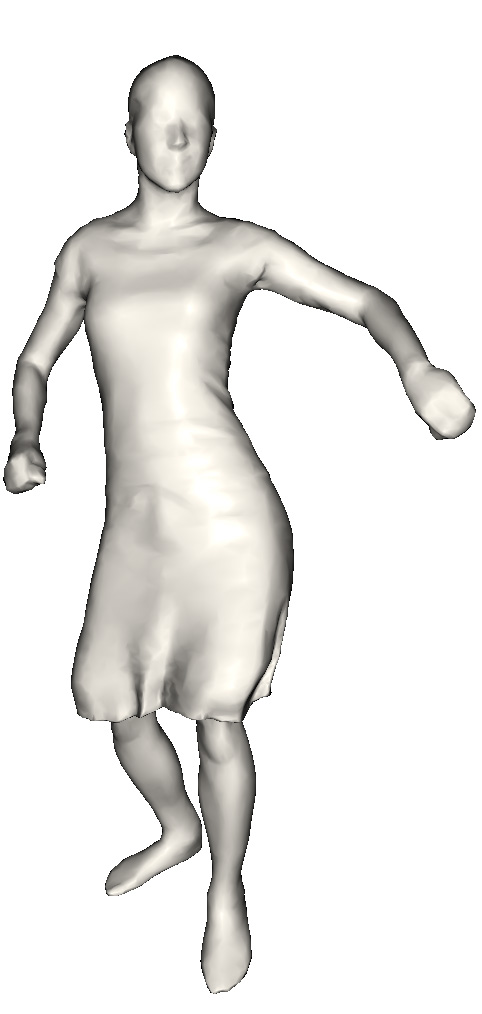}%
\includegraphics[height=\thispicheight]{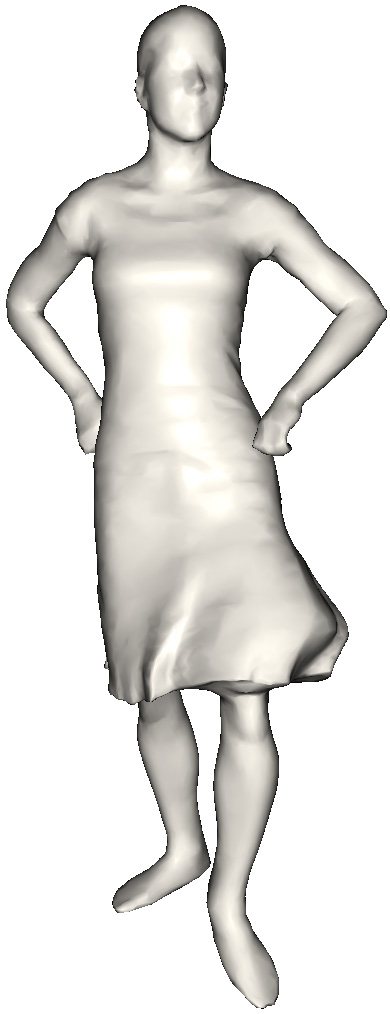}%
\hfill%
\includegraphics[width=\thispicwidth]{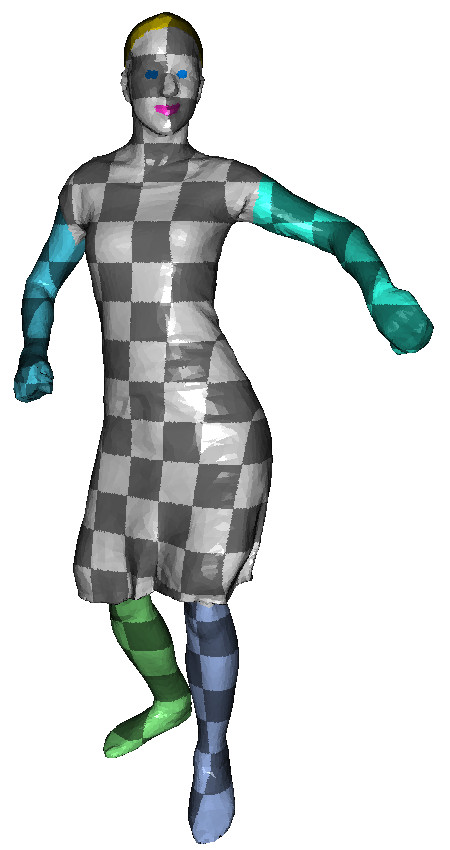}%
\includegraphics[width=.83\thispicwidth]{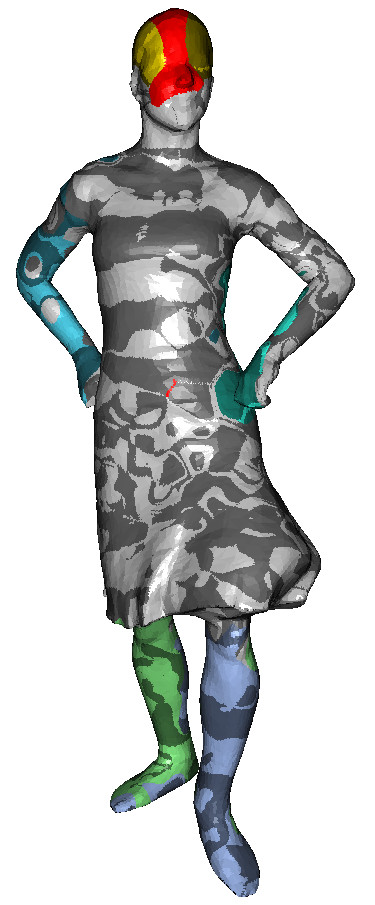}%
\hfill%
\includegraphics[width=\thispicwidth]{samba_clean_bim_result_highres_viz_source_front}%
\includegraphics[width=.83\thispicwidth]{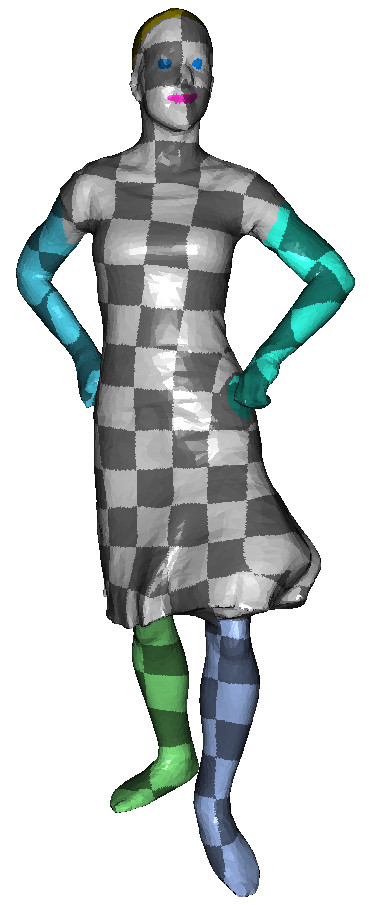}%
\hfill%
\includegraphics[height=\thispicheight]{samba_clean_bim_result_highres_viz_source_front}%
\includegraphics[width=.83\thispicwidth]{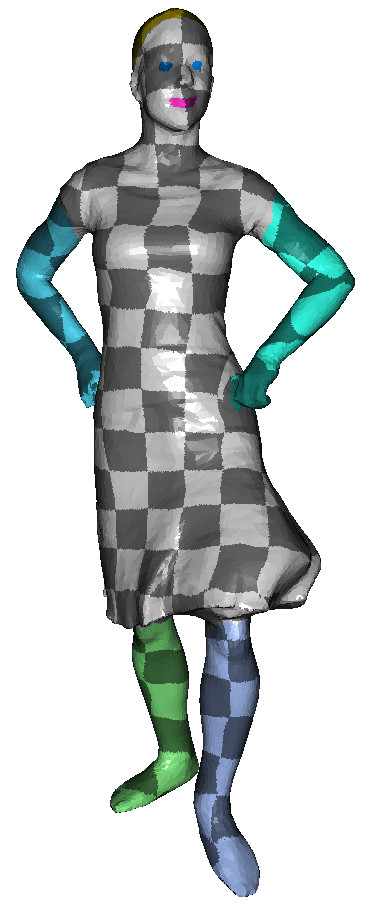}%
\\%New Line
{\small\sffamily%
\makebox[\fourpicwidth]{source and target}%
\hfill%
\makebox[\fourpicwidth]{Heat kernel maps \cite{Ovsjanikov2010}}%
\hfill%
\makebox[\fourpicwidth]{Blended intrinsic maps \cite{Kim2011}}%
\hfill%
\makebox[\fourpicwidth]{This paper}%
}%
\caption{Comparison to HKM and BIM on models with ground truth. For error rates see Figure \ref{fig_samba_plot}. Red indicates unmatched area; all further colors and the checker board have been painted on the source surface in order to visualize correspondences.}%
\label{fig_samba_clean}%
\end{figure*}

\begin{figure}%
\centering%
\includegraphics[width=0.9\columnwidth]{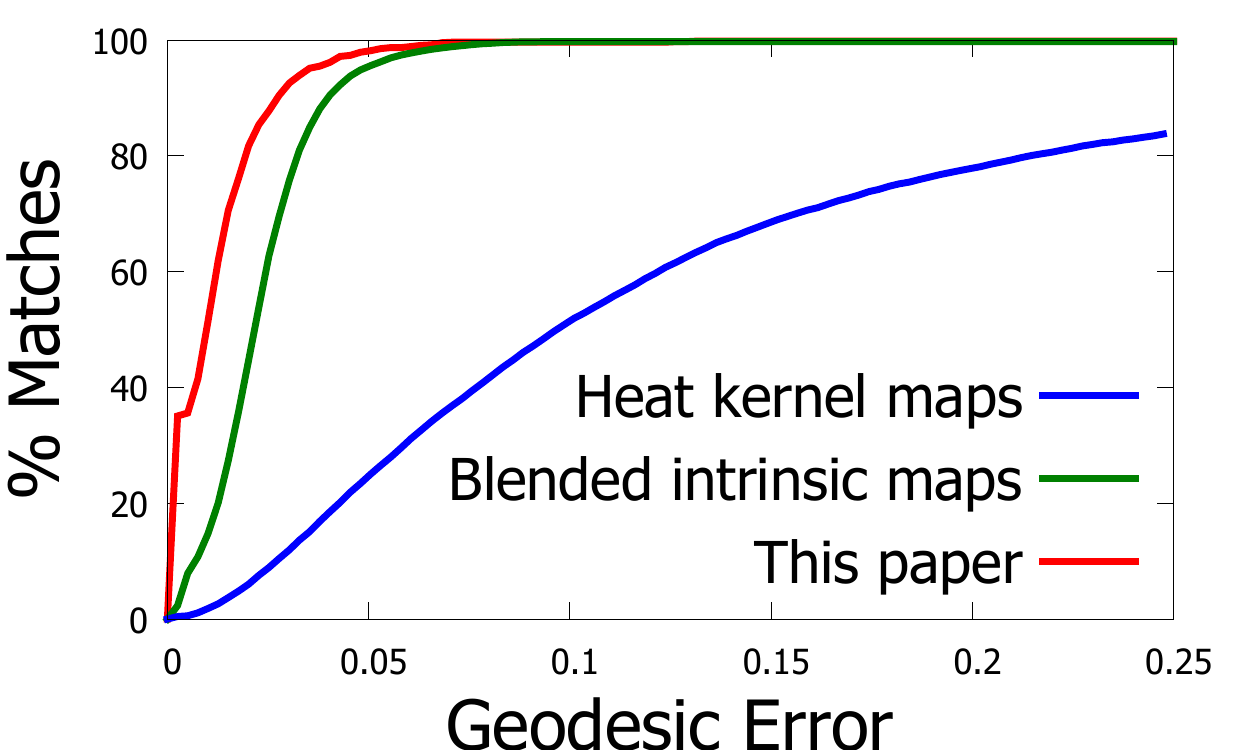}%
\caption{Cumulative error distributions for BIM, HKM, and our method based on the known ground truth. The models are shown in Figure \ref{fig_samba_clean}.}%
\label{fig_samba_plot}%
\end{figure}

We compare our method to HKM and BIM using the processed frames of the samba sequence. Figure~\ref{fig_samba_clean} shows the results. Note that while HKM leads to a result with visual artifacts, the results of BIM and our method are visually pleasing. Furthermore, since we have ground truth correspondences, we show the cumulative error distributions for all three methods in Figure \ref{fig_samba_plot}. Here, geodesic error is measured as a fraction of the square root of the surface area of $\surfaceT$. Note that our method numerically outperforms the two other methods. 

\resetlength{\thispicwidth}
\addtolength{\thispicwidth}{2cm}

The experiments conducted so far have shown that HKM, while being based on a solid theoretical foundation, leads to results of low quality when the aim is to find dense correspondences in datasets that contain noise and non-isometric distortion. Hence, in the following, we exclude this method from our comparisons.

\begin{figure}%
\centering%
\includegraphics[width=0.9\columnwidth]{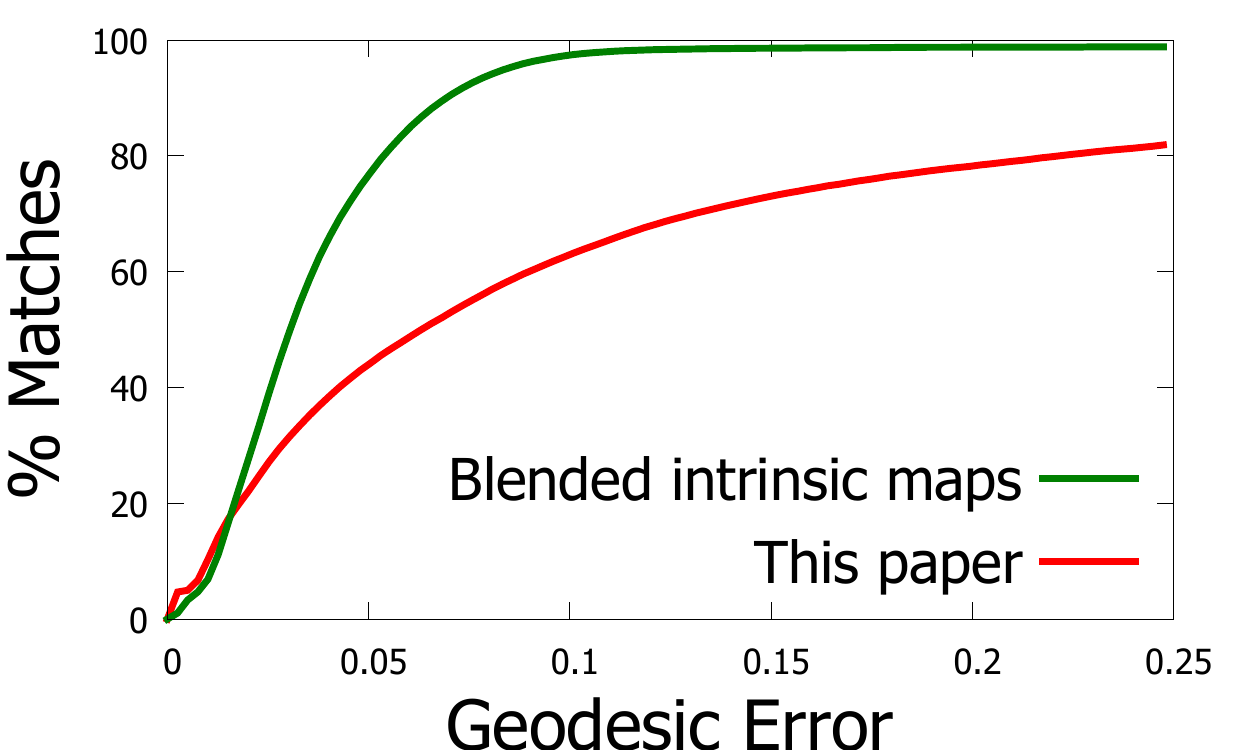}%
\caption{Cumulative error distributions for BIM and our method based on the known ground truth for SCAPE dataset.}%
\label{fig_scape_plot}%
\end{figure}

Second, we test our algorithm on the SCAPE dataset~\cite{Anguelov2005} consisting of 71 scans of a male scanned in different postures. In our experiment, we match the neutral posture to all 70 remaining postures using BIM and our approach. The cumulative error distributions for all 70 mappings are shown in Figure \ref{fig_scape_plot}.
This data differs from the samba frames in that it is globally near-isometric, but contains local areas of high-distortion (at joints for example). For this reason it provides a different kind of near-isometric test, and poses a greater challenge for our method, which does not exploit global assumptions. This is reflected in our error curve being below that of BIM. We set $\rho=1.9$ in this experiment to reduce the influence of partial isometric patches that were thrown off by high local distortions.

\subsection{Raw Scan Data}

Finally, we consider raw scan data acquired using different acquisition systems. This type of data is noisy and incomplete, and is therefore significantly more challenging to match than the data used in the previous experiments. By using data from a variety of acquisition systems, we show our methods robustness to different types of acquisition noise. We also show our method's robustness to other acquisition artifacts that violate global isometry: large holes and contacts.

First, we use the same two frames of the samba sequence by Vlasic et al.~\cite{vlasic2008} that were used above. However, this time, we consider the geometry reconstructed by a space-carving algorithm rather than template fitting. In this case, the frames we choose have different topology (as the hands merge with the body at the hips in one of the models) and are severely corrupted by noise. For this reason, we set $r_{min}=1.5R$, $r_{max}=3.4R$, and $\rho=1$. We use these models to compare our method to BIM, Tevs et al.'s method, and Sharma et al.'s method. Figure~\ref{fig_samba_contacts} shows the results. Note that the results using BIM and the method by Tevs et al. match parts of the body of $\surfaceS$ to the arms and legs of $\surfaceT$, thereby leading to visual artifacts. (See enlarged areas in Figure \ref{fig_samba_contacts}.) The method of Sharma et al., which is especially well suited to the scenario of handling contacts, leads to a result that covers the surfaces well. For this example, our method is run using two feature sets: first, using the standard features of our method and second, using the same image-based features used by Sharma et al. Our method produces a visually accurate mapping in both cases. However, when using standard features, the result of our method does not cover the right foot of the target surface, while the entire target surface is covered when using image-based features. Note that both the result by Sharma et al. and our results detect the contacts correctly and stop the growing in these regions. Hence, all areas of the surface are matched well. The improved performance with image-based features illustrates some of the technical challenges. The high level of surface noise makes matching geometric features difficult, which results in poorer sampling of the space of isometries, which in turn results in a poorer clustering result. We note however, that using the same features as Sharma et al., we obtain equal coverage and accuracy, and that when using purely geometric information, we outperform the other purely geometric methods tested.

\resetlength{\thispicwidth}
\addtolength{\thispicwidth}{2.0cm}

\begin{figure}
\centering
\includegraphics[width=\thispicwidth]{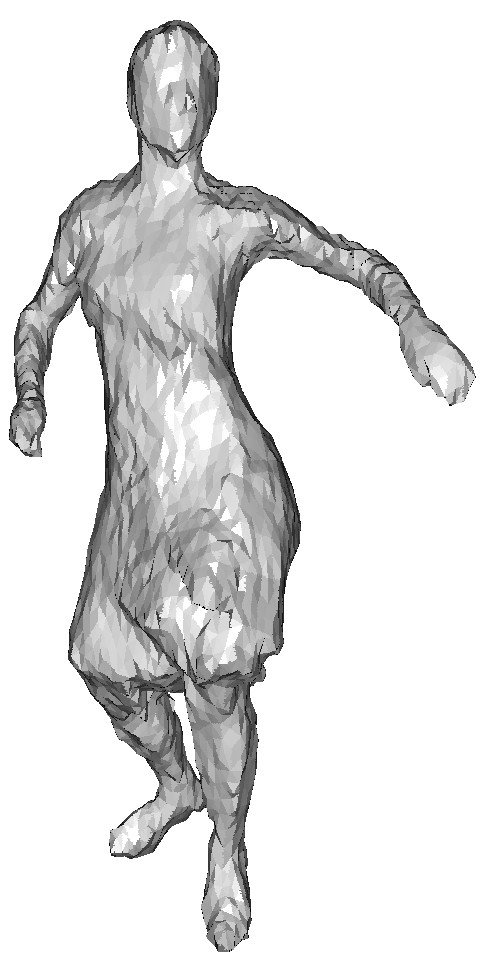}%
\hspace{1pt}%
\includegraphics[width=0.8\thispicwidth]{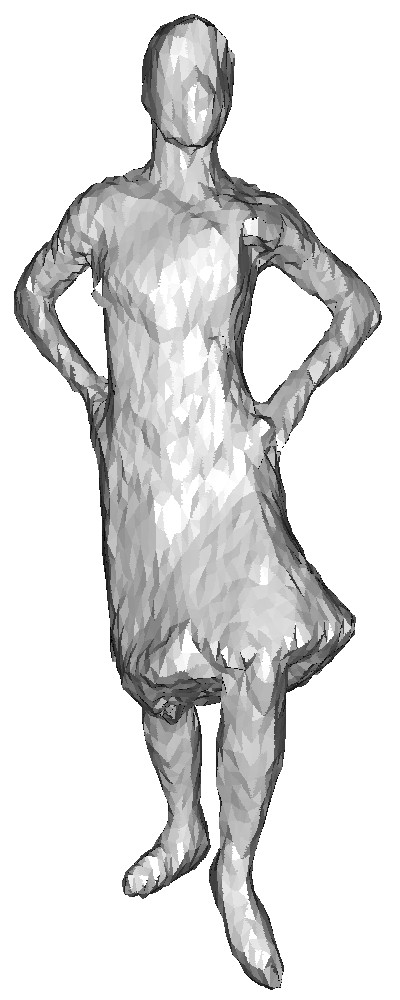}%
\hfill%
\includegraphics[width=1.95\thispicwidth]{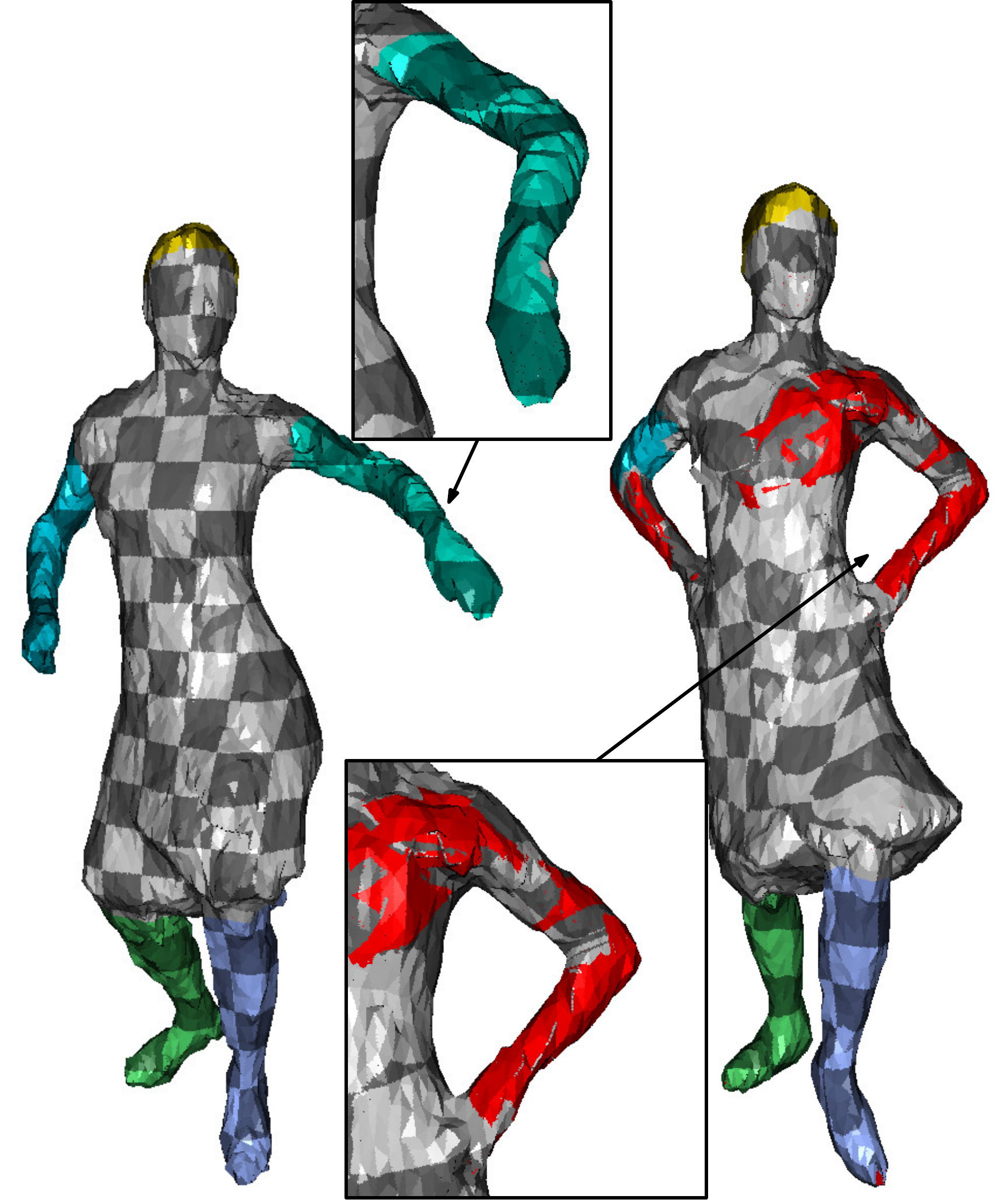}
\\%New Line
\vspace{0.1cm}
{\small\sffamily%
\parbox{1.8\thispicwidth}{\centering source and target}%
\hfill%
\parbox{1.8\thispicwidth}{\centering Blended intrinsic maps~\cite{Kim2011}}%
}%
\\%New Line
\vspace{0.2cm}
\includegraphics[width=1.95\thispicwidth]{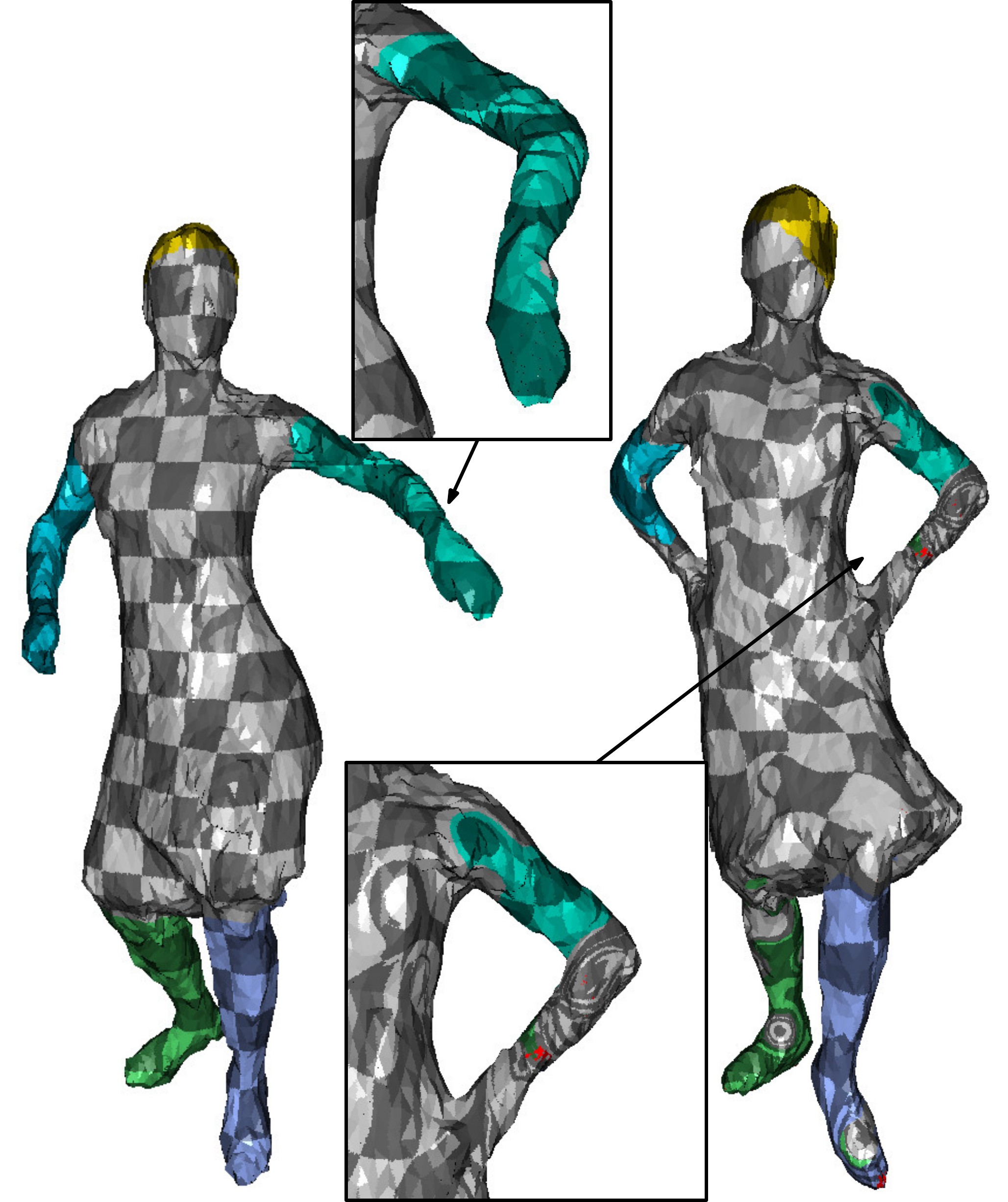}
\hfill%
\includegraphics[width=1.95\thispicwidth]{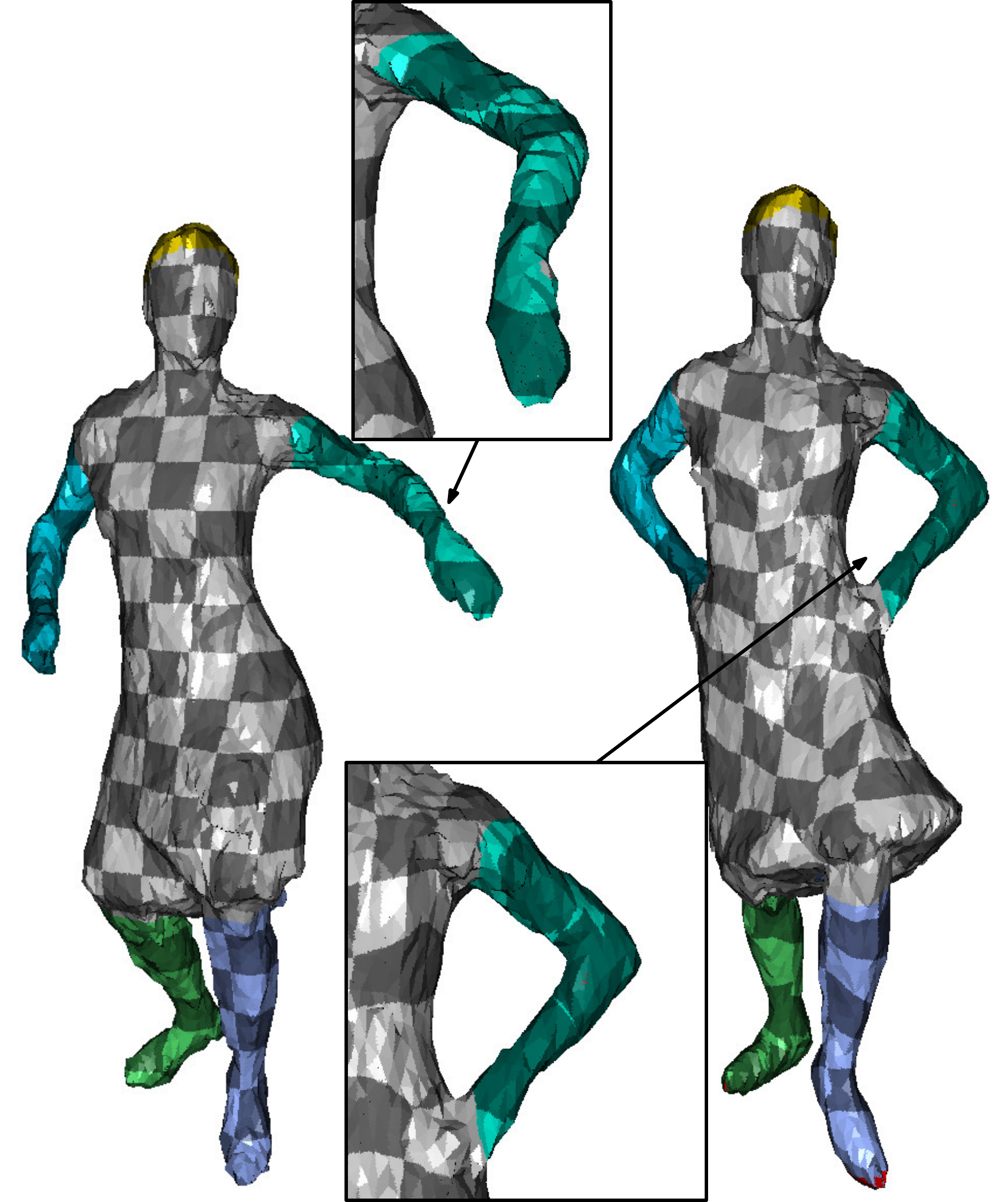}
\\%New Line
\vspace{0.1cm}
{\small\sffamily%
\parbox{1.8\thispicwidth}{\centering Tevs et al.~\cite{TevsAnimRec2012}}%
\hfill%
\parbox{1.8\thispicwidth}{\centering Sharma et al.~\cite{Sharma2011}}%
}%
\\%New Line
\vspace{0.2cm}
\includegraphics[width=1.95\thispicwidth]{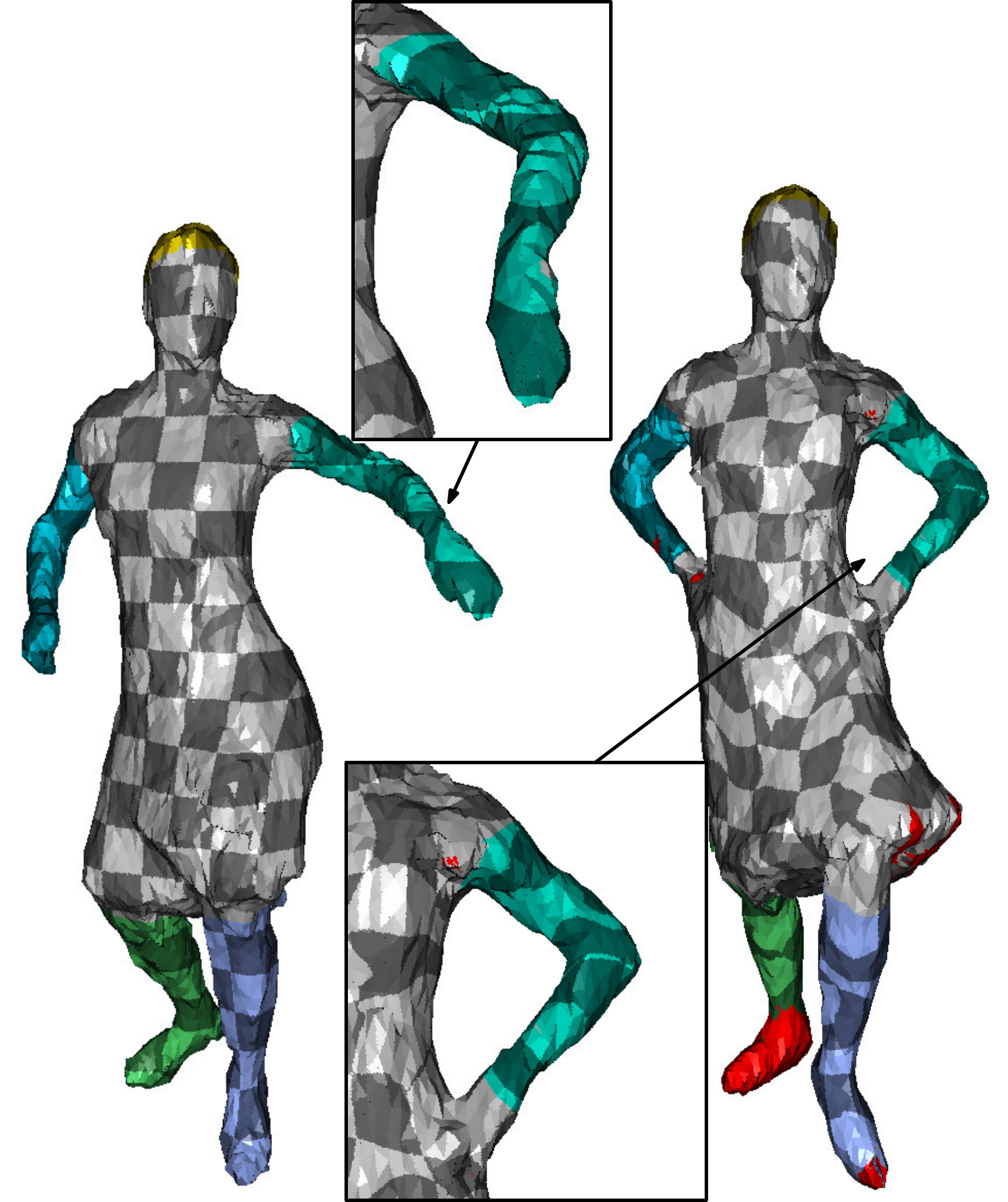}
\hfill%
\includegraphics[width=1.95\thispicwidth]{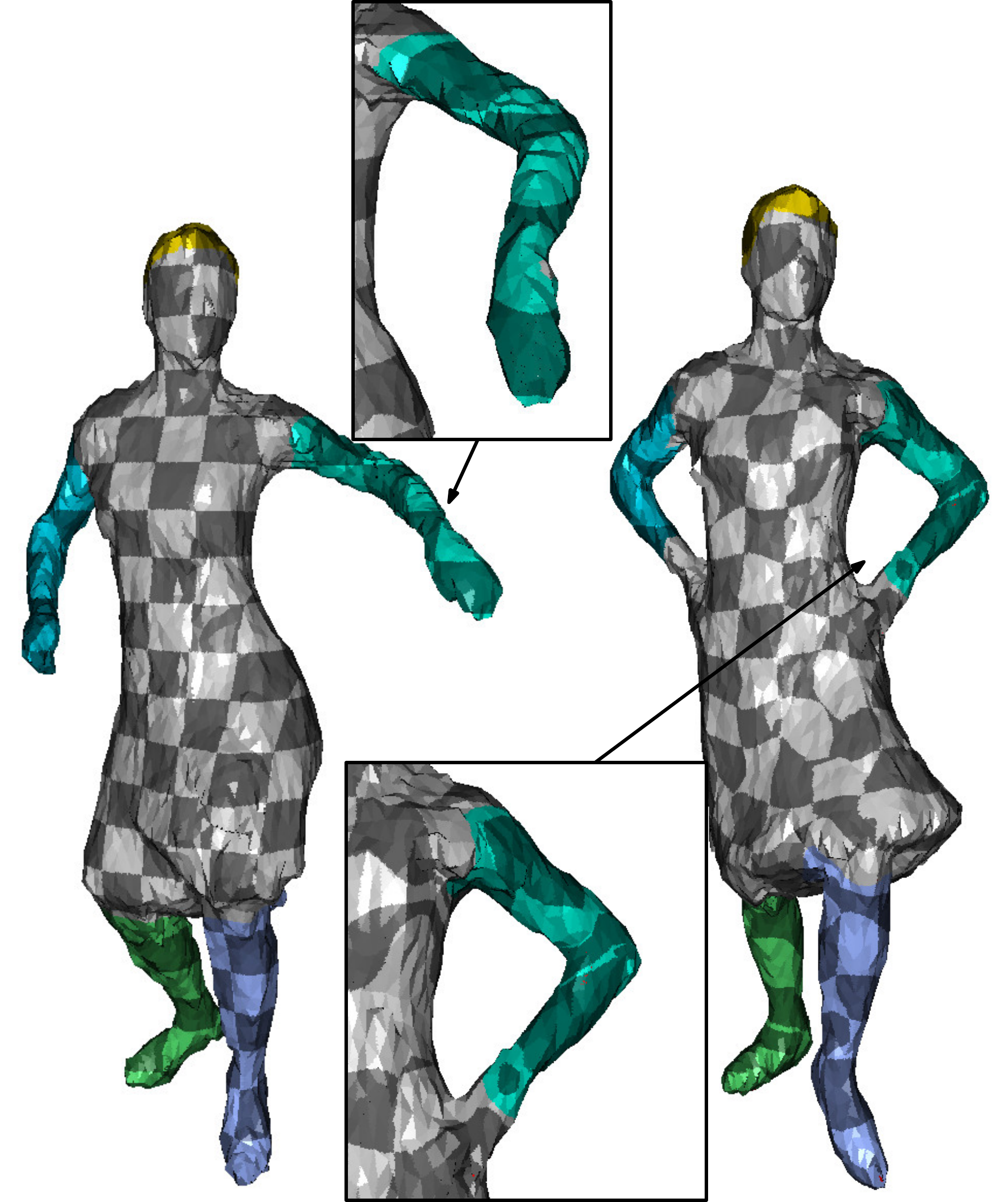}
\\%New Line
\vspace{0.1cm}
{\small\sffamily%
\parbox{1.8\thispicwidth}{\centering This paper}%
\hfill%
\parbox{1.8\thispicwidth}{\centering This paper with image-based features}%
}%
\caption{Comparison in presence of large contacts. Each pair shows $\surfaceS$ on the left and $\surfaceT$ on the right. Different data than in Figure \ref{fig_samba_clean}. Red again indicates unmatched area}
\label{fig_samba_contacts}
\end{figure}

\begin{figure}%
\centering
\includegraphics[width=\ltwopicwidth]{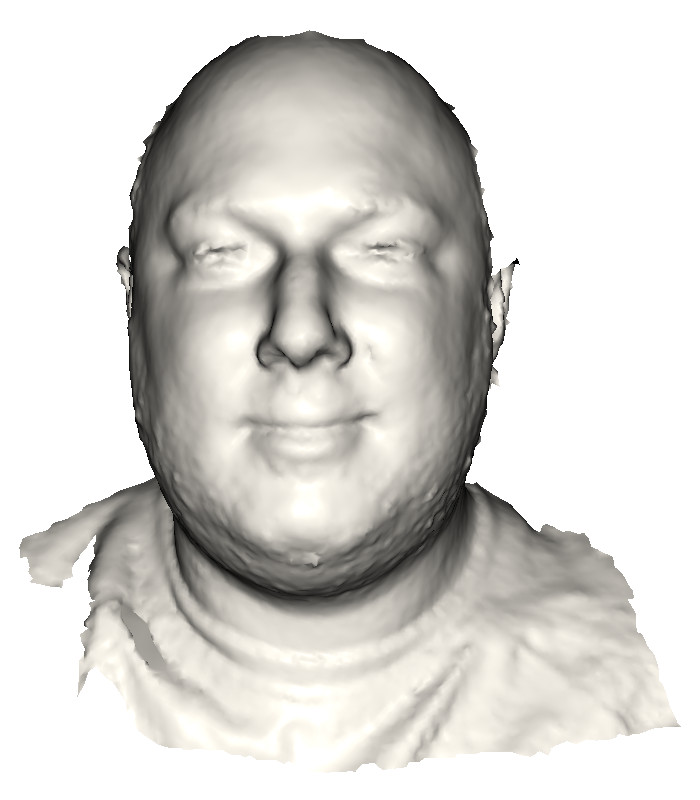}%
\hfill%
\includegraphics[width=\ltwopicwidth]{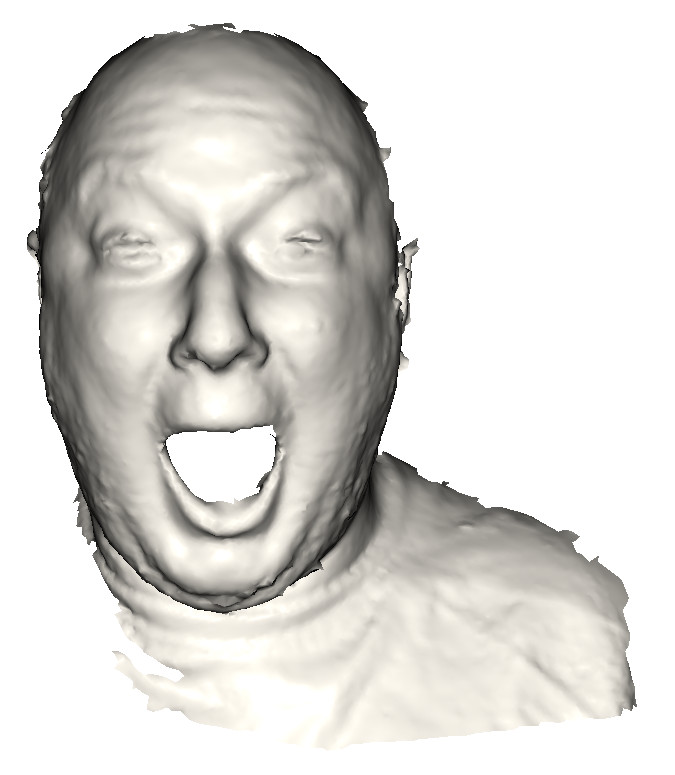}%
\\%New line
{\small\sffamily%
\makebox[\ltwopicwidth]{input: source (left) and target (right)}%
}%
\\%New line
\includegraphics[width=\ltwopicwidth]{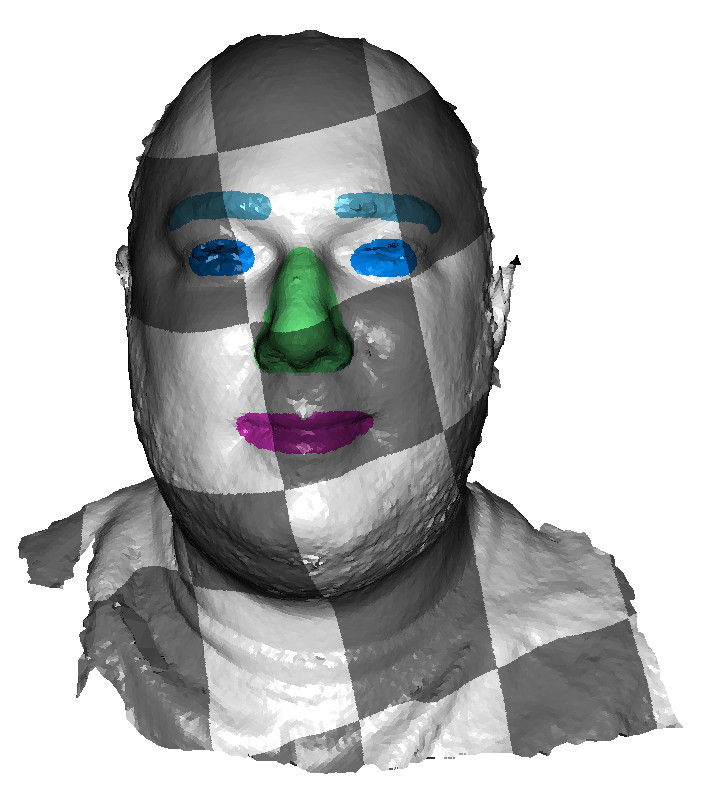}%
\hfill%
\includegraphics[width=\ltwopicwidth]{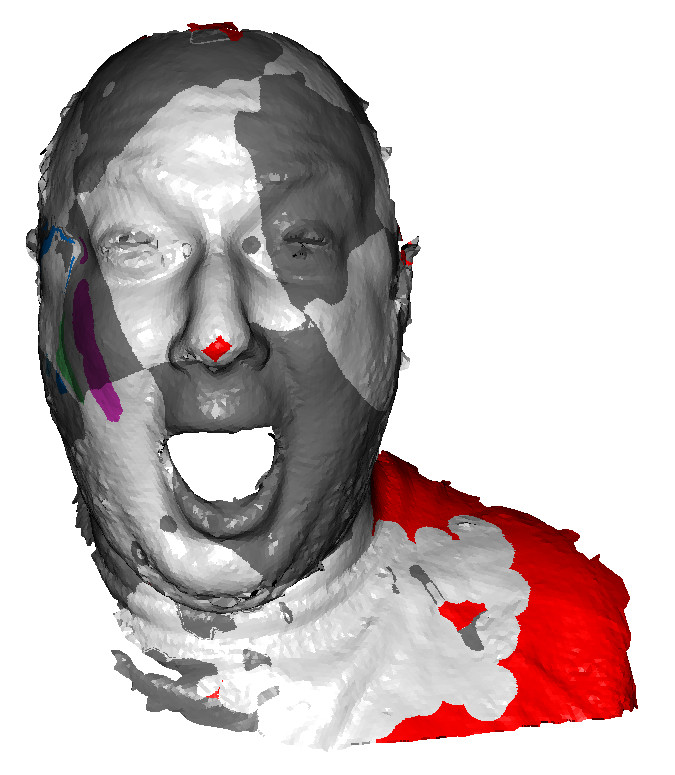}%
\\%New line
{\small\sffamily%
\makebox[\ltwopicwidth]{Blended intrinsic maps \cite{Kim2011}}%
}%
\\%New line
\includegraphics[width=\ltwopicwidth]{faces_bim_result_source_highresviz}%
\hfill%
\includegraphics[width=\ltwopicwidth]{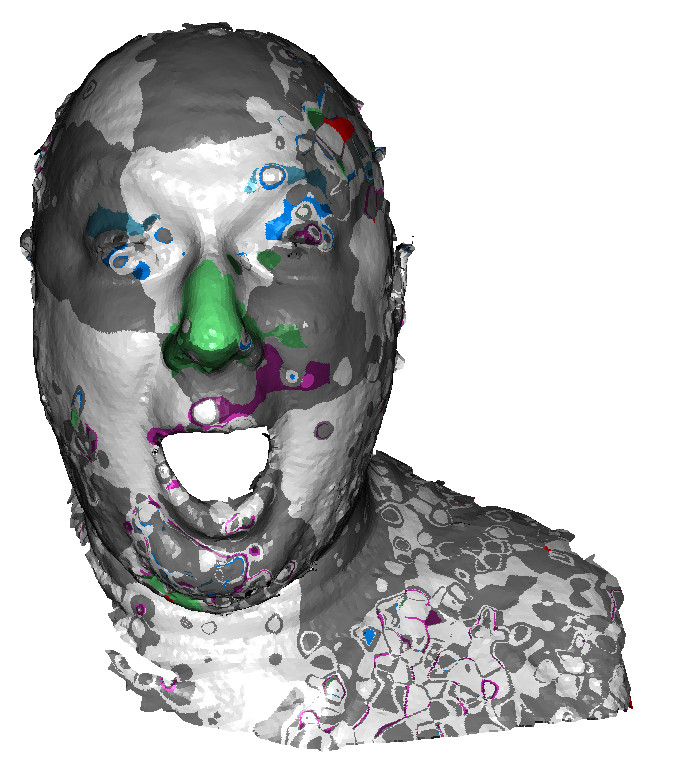}%
\\%New line
{\small\sffamily%
\makebox[\ltwopicwidth]{Tevs et al. \cite{TevsAnimRec2012}}%
}%
\\%New line
\includegraphics[width=\ltwopicwidth]{faces_bim_result_source_highresviz}%
\hfill%
\includegraphics[width=\ltwopicwidth]{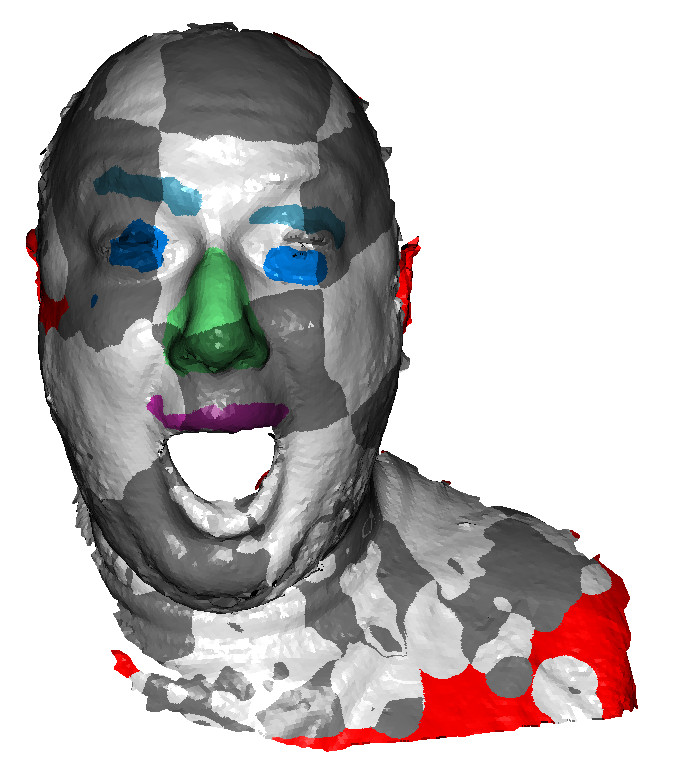}%
\\%New Line
{\small\sffamily%
\makebox[\ltwopicwidth]{This paper}%
}%
\caption{Comparison on data with significant topological changes and acquisition noise. Each pair shows $\surfaceS$ on the left and $\surfaceT$ on the right. Unmatched area marked in red.}%
\label{fig_face}%
\end{figure}

Second, we compare our method to BIM and Tevs et al.'s method using two models of the BU-3DFE face database~\cite{BU3DFE_2006}. The two models contain numerous small holes and outliers. Furthermore, the models have different topology because the mouth is closed in one model and open in the other one. For these reasons, we set $r_{min}=0.9R$, $r_{max}=1.7R$, and $\rho=1$. Figure~\ref{fig_face} shows the results. 
Note that our method obtains a mapping with higher visual accuracy than both BIM and the method of Tevs et al. In the case of BIM, this is to be expected, since the topological change breaks the global isometry. (Note the lips mapped to the side of the face.) The method of Tevs et al. produces a much more accurate result, but with still significant overall distortion and outliers (speckle-like effect). Our method leverages local information to fix partial isometries, and produces a mapping that largely preserves the semantic coloring with significantly lower distortion and without outliers.

\begin{figure}%
\centering
\includegraphics[width=0.97\ltwopicwidth]{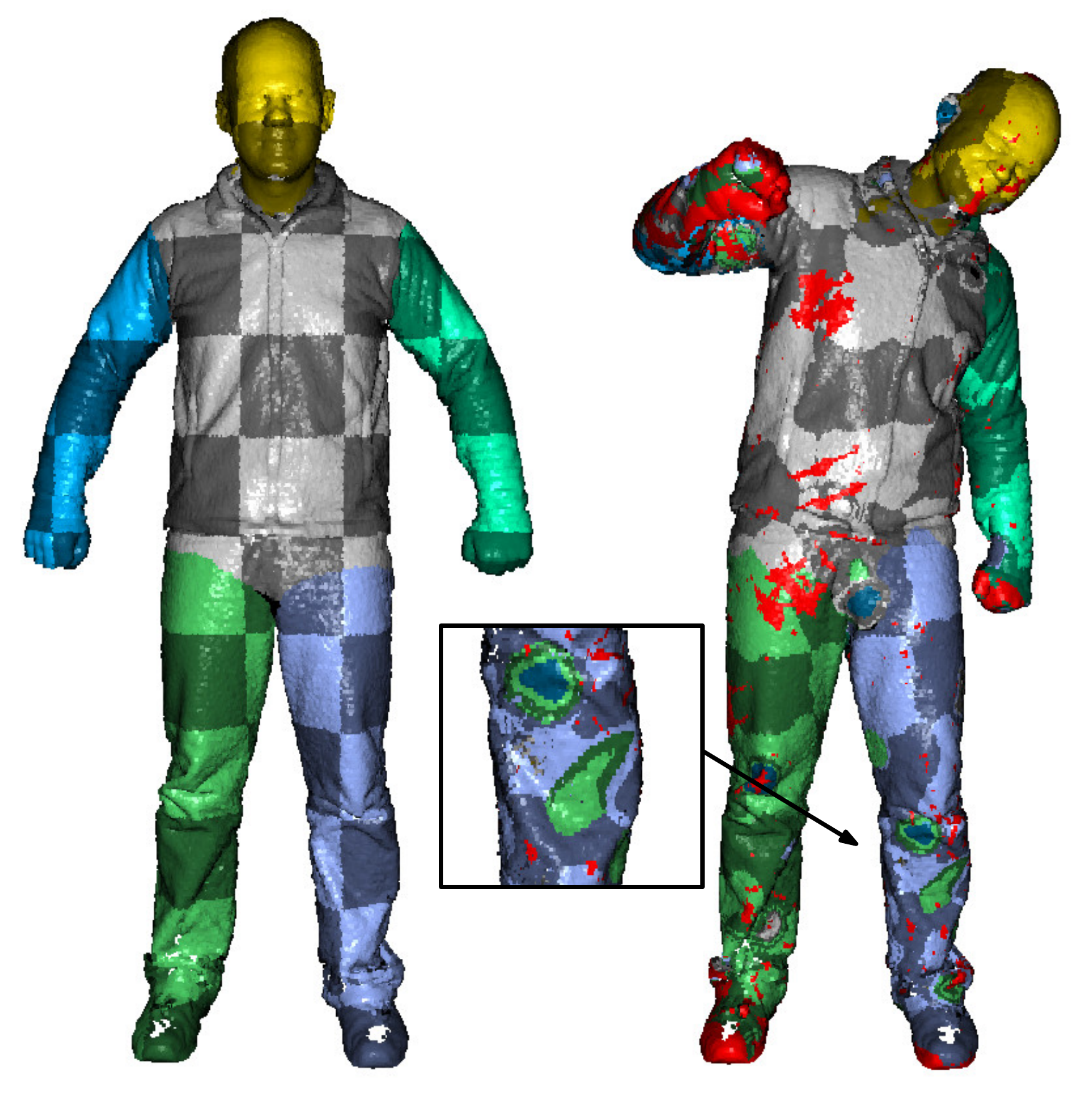}
\hfill
\includegraphics[width=\ltwopicwidth]{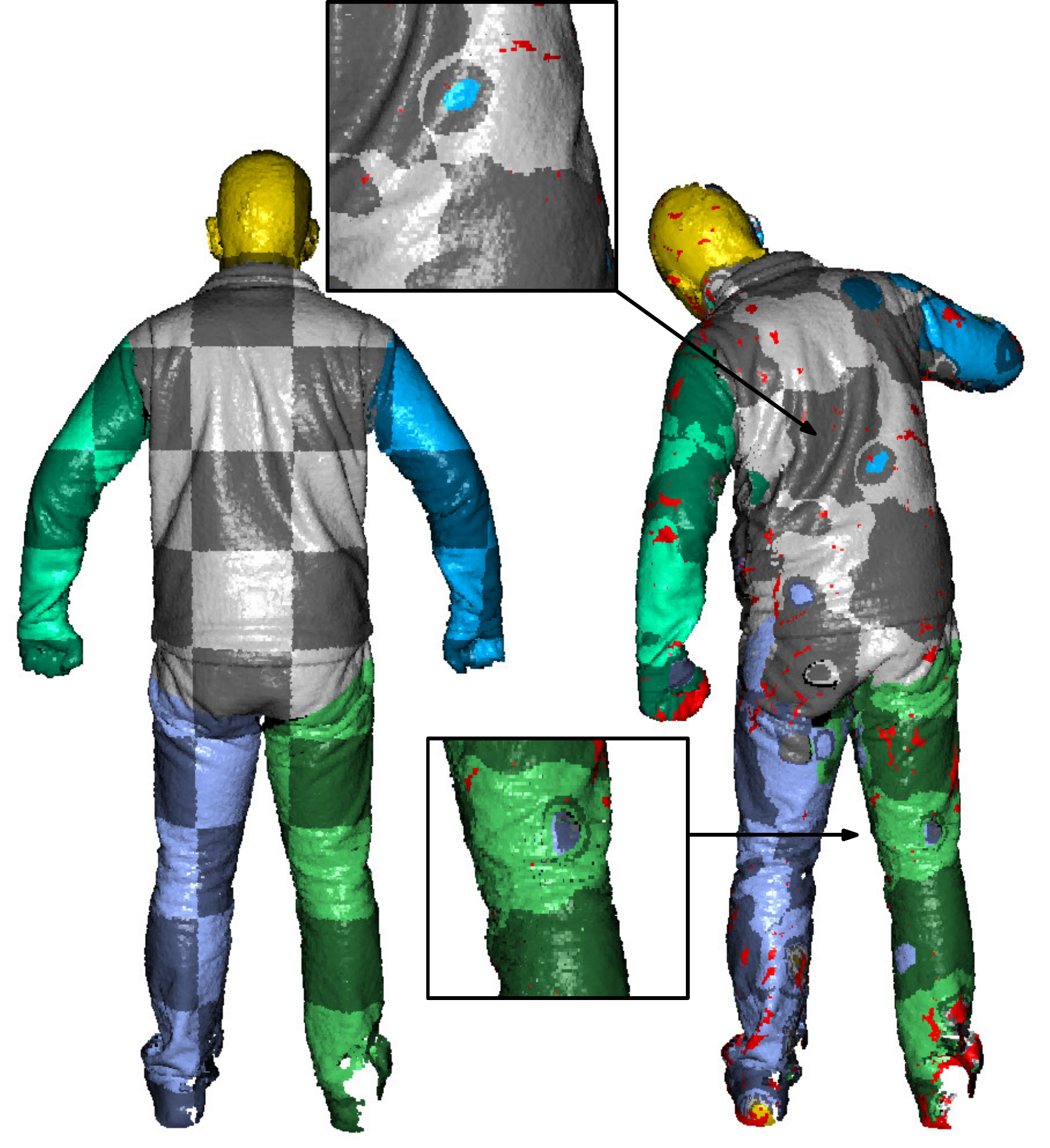}
\\
{\small\sffamily%
\makebox[\ltwopicwidth]{front}%
}%
\hfill%
{\small\sffamily%
\makebox[\ltwopicwidth]{back}%
}%
\\
{\small\sffamily%
\makebox[\width]{Tevs et al. \cite{TevsAnimRec2012}}%
}%
\\%New line
\includegraphics[width=0.97\ltwopicwidth]{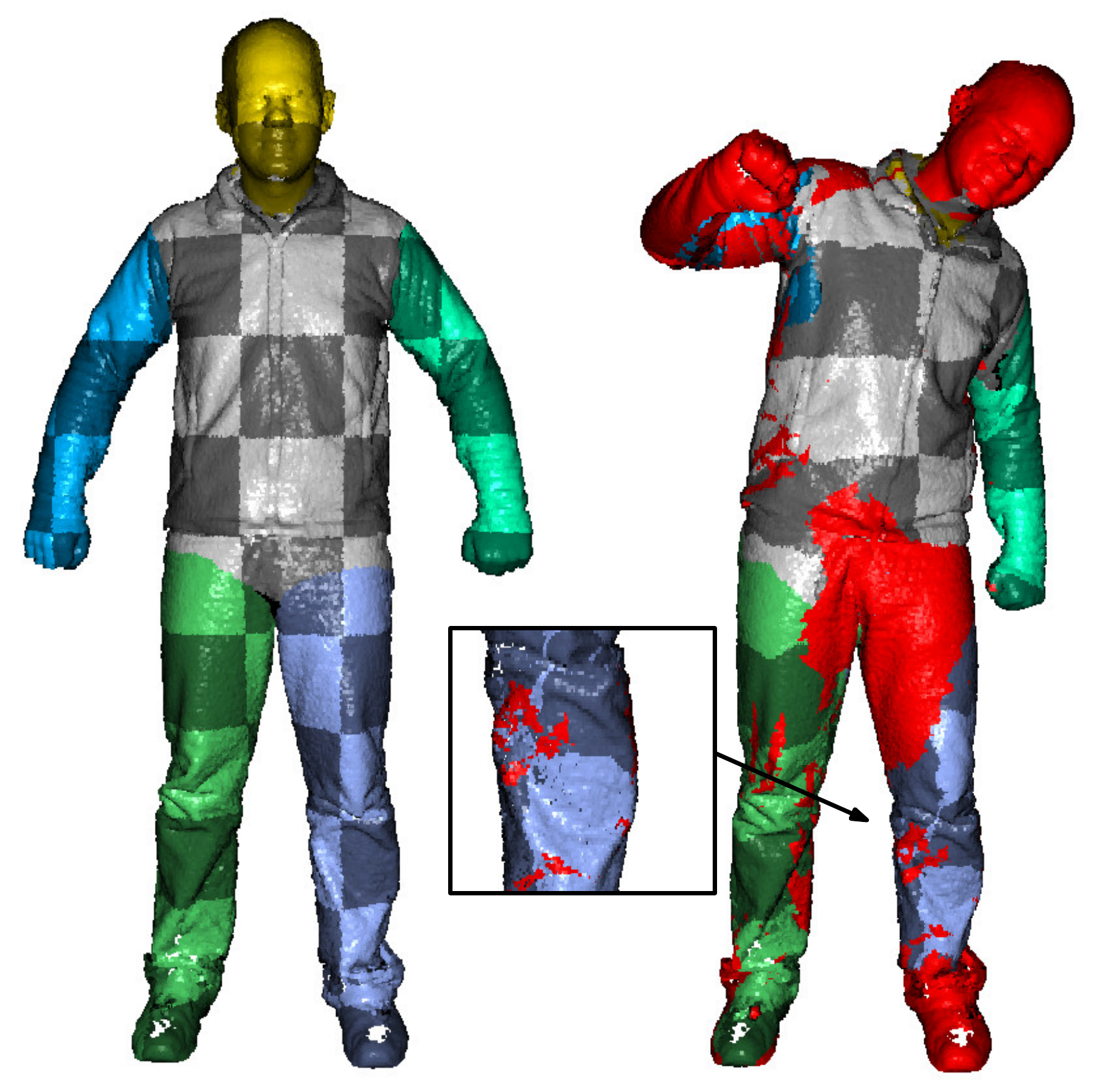}
\hfill
\includegraphics[width=\ltwopicwidth]{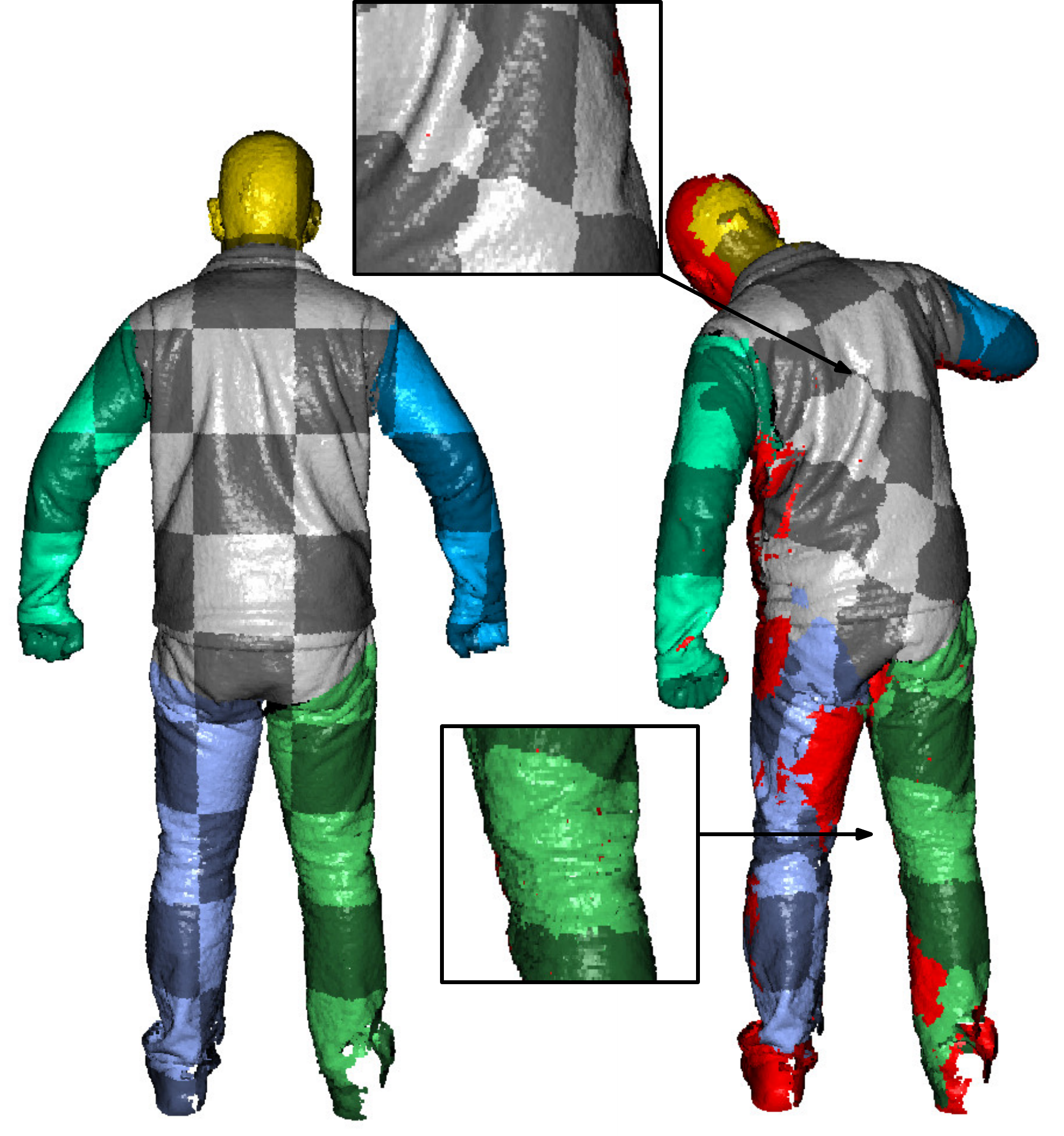}
\\
{\small\sffamily%
\makebox[\ltwopicwidth]{front}%
}%
\hfill%
{\small\sffamily%
\makebox[\ltwopicwidth]{back}%
}%
\\
{\small\sffamily%
\makebox[\width]{This paper}%
}%
\caption{Comparison on data with significant holes and acquisition noise. The first row shows the method of Tevs et al., while the second row shows our method. Within each row, the front and back are shown, in each case with $\surfaceS$ on the left and $\surfaceT$ on the right. Unmatched area shown in red.}%
\label{fig_carsten}%
\end{figure}

Third, we compare our method to that of Tevs et al.~on two point clouds acquired using a laser scanner. The two models contain numerous holes and outliers. As the models are point clouds and BIM requires input meshes, we do not compare our result to BIM for this experiment. For these models, we set $r_{min}=1.5R$, $r_{max}=3.4R$, and $\rho=1$. The results are shown in Figure \ref{fig_carsten}. As can be seen, the method of Tevs et al.~obtains better coverage, however our method has fewer outliers--islands of incorrectly mapped points within larger smoothly mapped regions (see the enlarged parts of Figure \ref{fig_carsten}). The suboptimal coverage of our method is due to the difficulty in matching features on surfaces plagued by missing data. A feature descriptor less sensitive to holes could improve this.

\subsection{Limitations}

\begin{figure}%
\centering%
\includegraphics[height=4.0cm]{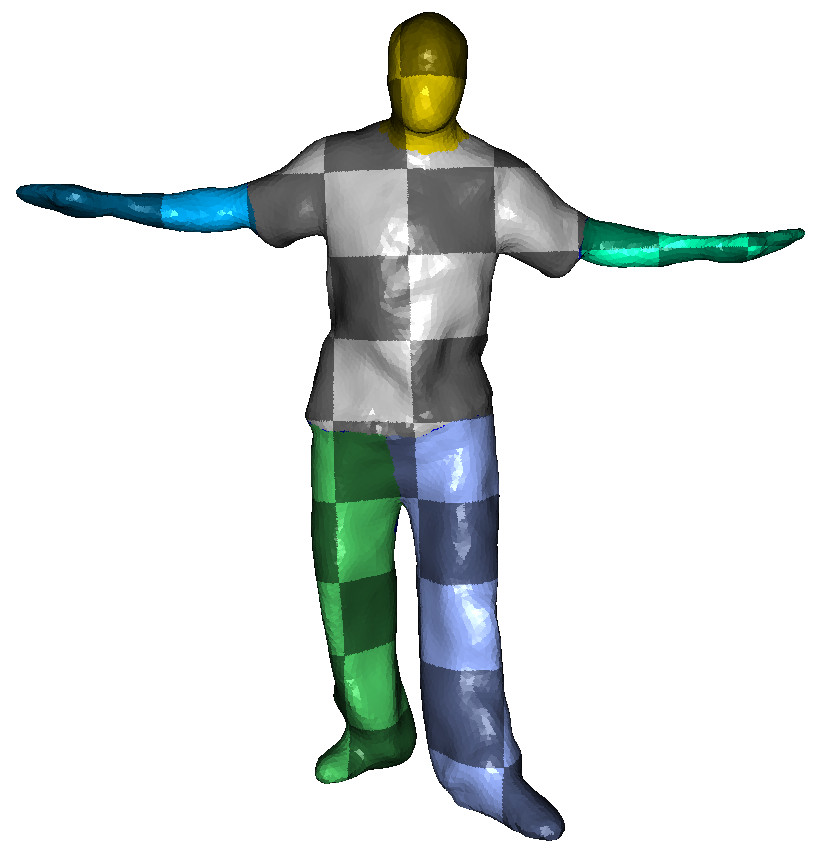}%
\hfill%
\includegraphics[height=4.0cm]{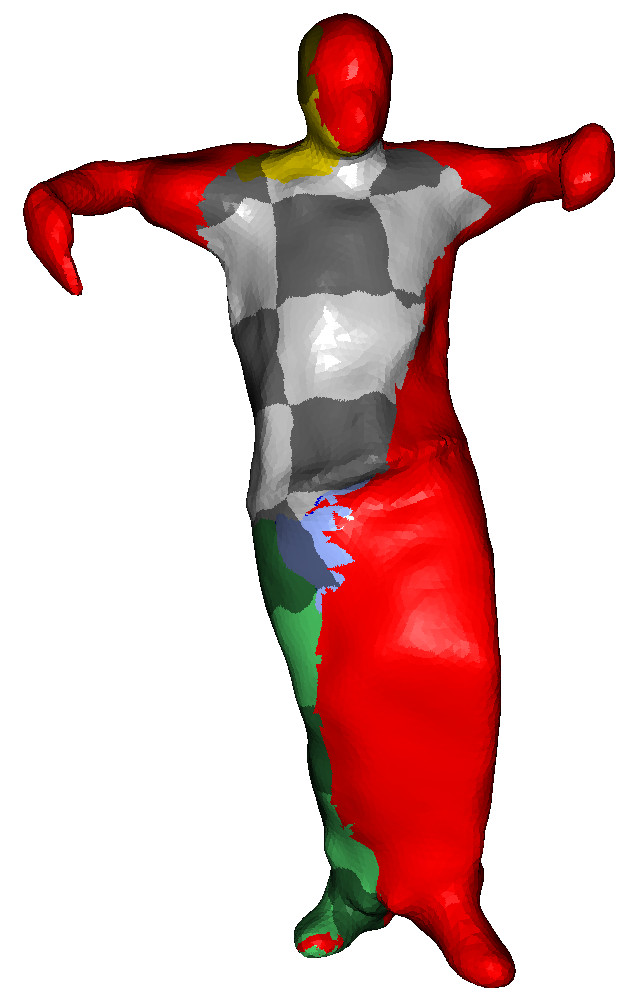}%
\\[\baselineskip]%
\includegraphics[height=4.0cm]{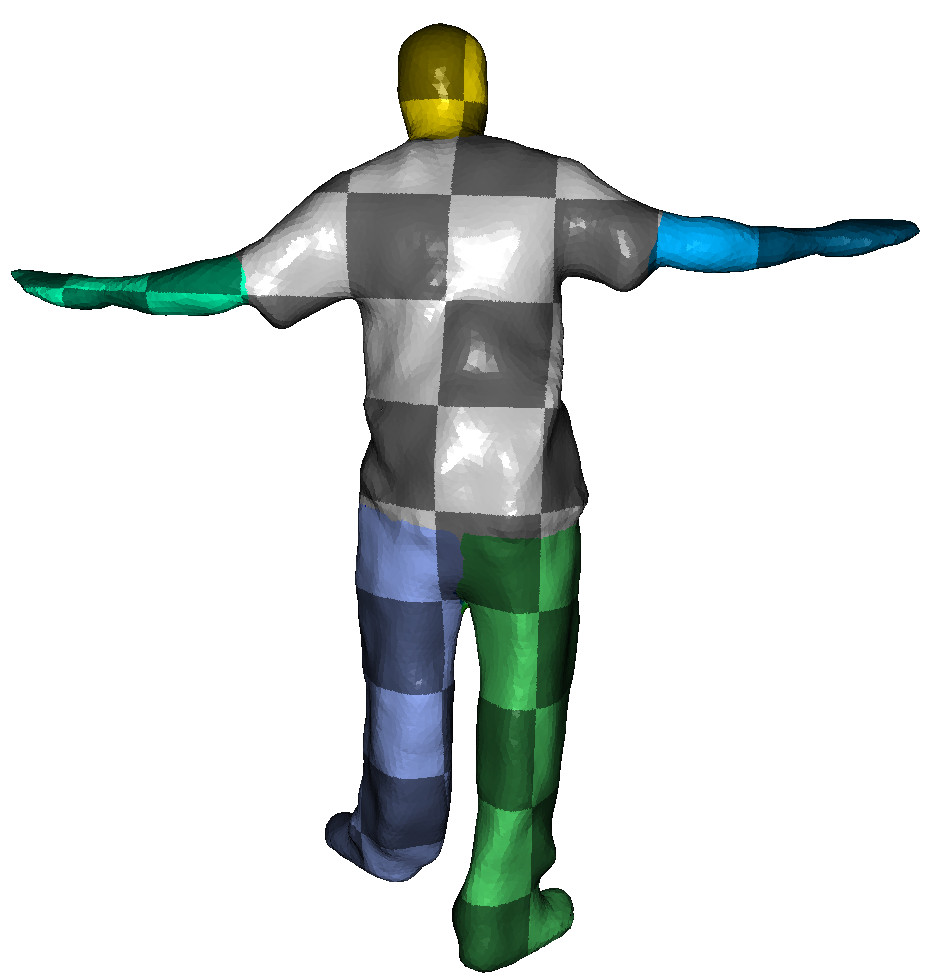}%
\hfill%
\includegraphics[height=4.0cm]{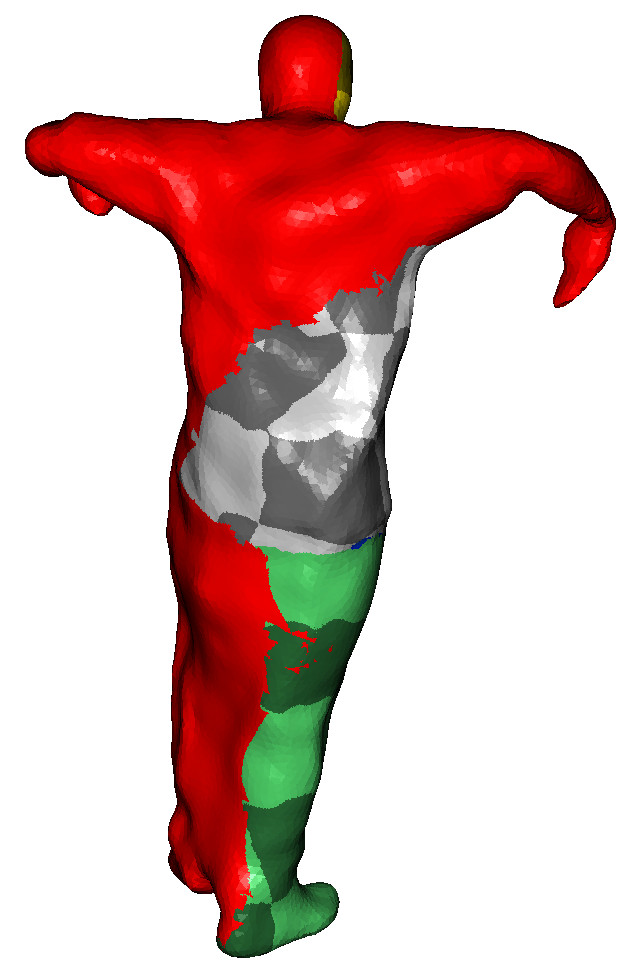}%
\caption{A single partial mapping for an example with large topological changes. The red area is unmatched.}
\label{fig_flashkick}
\end{figure}

In the previous sections, we have demonstrated that our method not only has theoretical advantages, but also computes results that improve upon the state of the art results in challenging cases where the input data is a pair of raw scans with topological noise.

We now discuss some limitations of our algorithm. Our algorithm is based on growing near-isometric mappings between partial regions of two surfaces and then clustering consistent mappings together. In this way, our algorithm enumerates a set of near-isometric mappings. For models that exhibit a large number of partial intrinsic symmetries, this technique enumerates a large set of near-isometric mappings where many of the mappings are inconsistent with each other. Since we stop growing new near-isometric regions after the first 200 oriented feature point matches have been considered, for shapes with a large number of partial near-isometric symmetries, it may happen that there is no cluster of consistent mappings covering a large area of $\surfaceS$.

An example where the clustering step fails to identify a cluster of consistent mappings covering a large area of $\surfaceS$ is shown in Figure~\ref{fig_flashkick}. The two frames of the flashkick dataset~\cite{starck_2007} contain a large topological change due to a merge of the subject's pants in one of the models. Note that our algorithm correctly stops the growing of single partial mapping in this area, as shown in Figure~\ref{fig_flashkick}. However, since the legs and core of the target body are intrinsically symmetric (similar to cylindrical), many inconsistent partial mappings are found by the algorithm, and they cannot be clustered in a consistent way.

We should also note that the computational costs are quite high. This is partially due to unoptimized code, but an algorithmic shortcoming is the rather simple feature-matching algorithm for finding start positions and directions. Optimizing this was not the focus of this work; our method should rather be understood as an alternative for the dense matching step where it can replace previous approaches based on conformal maps~(\cite{Lipman2009,Kim2011} and follow-ups), heat-kernel maps (\cite{Ovsjanikov2010} and follow-ups), or geodesic triangulation with landmarks (\cite{Huang2008,Tevs2009} and follow-ups) in order to handle partial matching.

For future work, to address this limitation, we plan to combine our approach with an approach that detects continuous intrinsic symmetries~\cite{butscher_sgp2010}, thereby reducing the search space for the initial feature matching and allowing us to efficiently enumerate all partial matches. Similarly, our framework could be extended to find partial intrinsic symmetries within a single object.

\section{Conclusion}
\label{sec:conclusions}

We have analyzed the complexity of the isometric matching problem under global and local isometry assumptions and based on this analysis we have introduced a new approach to solve the partial isometric matching problem using a representation for partial isometries that is both low-dimensional and redundant. Underpinning this is the fundamental observation that isometric mappings can be determined using purely local information and have only three degrees of freedom on 2-manifolds. The local metric propagation algorithm we derived from this observation is designed to handle topological noise that could affect large portions of the model, including both large holes and contacts. The redundancy in the representation can be exploited to increase robustness of and to combine partial matches.%Furthermore, we discussed that the space of partial isometric mappings has a redundancy in the representation, which can be used to increase the robustness of the computed partial matches in practice. 
We have shown how a direct implementation of this theoretical framework can be used to match challenging surfaces with different types of topological noise.%We presented a direct non-optimized implementation of this method and demonstrated our method's ability to handle different types of large topological noise on noisy scan data.

The insights gained by studying the partial isometric matching problem have the potential to impact other shape processing tasks. For instance, the representation for partial isometric matches introduced in this paper can be used to derive new algorithms to detect partial symmetries of shapes. For future work, we plan to further investigate this option.

\section*{Acknowledgements}

The authors thank Art Tevs, Aurela Shehu, Waqar Khan and Avinash Sharma for their help in conducting the comparative evaluation, Vladimir Kim for making his code available, and Art Tevs, Silke Jansen, Alexander Berner, Qi-Xing Huang, and Leonidas Guibas for discussions. We also thank the anonymous reviewers for their valuable comments that we used to improve the paper.
 %We also thank Riemannian geometry for being so devoid of degrees of freedom with respect to metric-preserving mappings.


%-------------------------------------------------------------------------
\newpage

\end{document}